\documentclass[journal]{IEEEtran}
\usepackage{amsmath,amsfonts}
\usepackage{algorithm}
\usepackage{array}
\usepackage{textcomp}
\usepackage{stfloats}
\usepackage{float}
\usepackage{url}
\usepackage{verbatim}
\def\BibTeX{{\rm B\kern-.05em{\sc i\kern-.025em b}\kern-.08em
    T\kern-.1667em\lower.7ex\hbox{E}\kern-.125emX}}
\usepackage{balance}

\usepackage{multirow}
\usepackage{booktabs}
\usepackage{tabularx}
\usepackage{xcolor}
\usepackage{cite}
\usepackage{xurl}
\usepackage{hyperref}
\usepackage{graphicx} 
\usepackage{subcaption}
\usepackage{algpseudocode}
\algrenewcommand\algorithmiccomment[1]{\hfill // #1} % for comment in sudo code
\usepackage{orcidlink} % provides \orcidlink{} command

\begin{document}

\bstctlcite{IEEEexample:BSTcontrol}

\title{Unsupervised Detection of Entry and Exit Regions from Vehicle Trajectories for Camera-Agnostic Turning Movement Counts}

\author{
Parikshit~Singh~Rathore\textsuperscript{1,2,*}\orcidlink{0009-0000-8022-9195},
Vishwajeet~Pattanaik\textsuperscript{2,*}\orcidlink{0000-0002-7389-5955}
and
Punit~Rathore\textsuperscript{1,2}~\orcidlink{0000-0003-4835-4556}\\
\textsuperscript{1}Robert Bosch Centre for Cyber-Physical Systems (RBCCPS), \\
Indian Institute of Science (IISc), Bengaluru, India \\
\textsuperscript{2}Centre for infrastructure, Sustainable Transportation and Urban Planning (CiSTUP), \\
Indian Institute of Science (IISc), Bengaluru, India \\
\{parikshits, vishwajeetp, prathore\}@iisc.ac.in
\thanks{\textsuperscript{*}Both authors contributed equally to this work.}
}
        % <-this % stops a space



\maketitle

\begin{abstract}
Turning movement counts are essential for intersection-level traffic management, yet their collection remains predominantly manual due to the cost of per-camera region annotation. This paper presents an unsupervised pipeline that identifies entry and exit regions directly from raw vehicle trajectories extracted via object detection and multi-object tracking, requiring no manual annotation, camera calibration, or prior knowledge of intersection geometry. Unlike trajectory clustering methods that classify individual trajectories using pairwise similarity and must be re-executed on every new batch, the proposed pipeline clusters initial and terminal point locations to produce persistent spatial region polygons that classify future trajectories by point-in-polygon containment at linear cost. The pipeline comprises six sequential steps, five of which introduce configurable parameters evaluated through a systematic statistical analysis spanning 17,100 pipeline executions across 9 surveillance cameras capturing dense heterogeneous traffic in Bengaluru, India, and 10 sequences from the UA-DETRAC benchmark dataset. Both parametric and nonparametric testing frameworks identify three consistently significant parameters and yield an empirically grounded recommended configuration. Under this configuration, the pipeline achieves a median classification error of 3.4\% across all 25 Bengaluru cameras, including 16 held-out locations, with a median per-turning-movement GEH of 2.43. Compared with two trajectory clustering baselines, the proposed pipeline exhibits greater stability across camera views and lower computational cost, at the expense of higher median error. Extended evaluation demonstrates that calibration clips of at least 60 minutes and peak-traffic selection further improve region estimation quality.
\end{abstract}

\begin{IEEEkeywords}
% ref keyword policy: https://ieee-itss.org/pub/t-its/#:~:text=Grammatically%20correct.-,Keywords,-index%20terms%20must
Turning pattern, Transportation planning and design, Trajectories, Traffic networks, Smart cities, Computer vision
\end{IEEEkeywords}

%%%%%%%%%%%%%%%%%%%%%%%%%%%%%%%%%%%%%%%%%%%%%%%%%%%%%%%%%%%%%%%%%%%%%%%%%%%%%%%%%%%%%%%%%%%%%%%%%%%%
%%%%%%%%%%%%%%%%%%%%%%%%%%%%%%%%%%%%%%%%%%%%%%%%%%%%%%%%%%%%%%%%%%%%%%%%%%%%%%%%%%%%%%%%%%%%%%%%%%%%
\section{Introduction}
\label{sec:intro}

Turning movement counts (TMCs) classify vehicles by their entry and exit paths at intersections and are among the most critical inputs for urban traffic management. Accurate TMCs underpin signal timing, capacity planning, and safety analysis~\cite{Tituana2022}. Yet TMC data collection remains dominated by manual methods originating in the 1950s. The 2012 U.S.\ National Traffic Signal Report Card assigned a failing grade to traffic monitoring practices nationwide~\cite{NTOC2012}, and TMC data is typically collected for only one to two days per year per intersection due to prohibitive labor costs~\cite{Day2010}. The resulting data scarcity limits the ability of transportation agencies to respond to evolving traffic patterns, particularly in rapidly growing cities where intersection demand changes faster than manual surveys can track.

The growing deployment of surveillance cameras in metropolitan areas worldwide presents an opportunity to close this data gap, though not through purpose-built traffic infrastructure. India's Safe City project, for instance, is deploying tens of thousands of cameras across eight metropolitan cities for public safety rather than traffic monitoring. Bengaluru alone operates over 7{,}000 such cameras at junctions, markets, and transport hubs~\cite{IE2023}, while Delhi is expected to deploy approximately 10{,}000 under the same programme~\cite{IE2025}. Similar surveillance expansions are underway across Southeast Asia, the Middle East, and Africa, driven by public safety mandates and rapid urbanization. These cameras frequently encompass intersections and high-volume road segments within their fields of view, yet they are neither positioned nor calibrated for traffic engineering; deploying dedicated traffic monitoring infrastructure at comparable scale remains prohibitively expensive, even for high-income nations~\cite{Saldivar-Carranza2021}. Methods that can extract traffic engineering value from existing surveillance networks, without requiring per-camera calibration or manual annotation, therefore address a practical need that extends well beyond any single city or country. The requirements that any such method be unsupervised and camera-agnostic are not merely desirable technical properties; they are constraints imposed by the deployment reality of networks comprising thousands of uncalibrated, heterogeneously mounted cameras that cannot feasibly be configured individually.

Vehicle trajectories extracted from video through object detection~\cite{Zou2019} and multi-object tracking~\cite{Yilmaz2006} encode precisely the information needed for TMC: where each vehicle entered the field of view and where it exited. The prevailing approach to obtaining TMCs from trajectories, however, still involves manually delineating entry and exit regions and subsequently mapping trajectories through them~\cite{Shirazi2016, Gloudemans2021, Pakdamansavoji2024}. As Saldivar-Carranza et al.~\cite{Saldivar-Carranza2021} observe, this per-intersection manual specification constitutes the primary bottleneck preventing network-wide deployment. The present work addresses this bottleneck.

Automating the identification of entry and exit regions from observed trajectories is complicated by several compounding factors: perspective distortion from uncalibrated cameras~\cite{Solh2009}; diverse intersection geometries that preclude a single spatial template; heterogeneous, non-lane-following traffic that creates two-dimensional flow patterns invalidating lane-based assumptions~\cite{Mallikarjuna2009}; and operational variability across camera networks in mounting height, focal length, and environmental conditions. These challenges, and the extent to which prior work has addressed them, are reviewed in Section~\ref{sec:related_work}.

We propose an unsupervised pipeline that identifies entry and exit regions directly from raw vehicle trajectories. Unlike trajectory clustering methods, which classify individual trajectories by movement type using pairwise similarity measures and must be re-executed on every new batch, the proposed pipeline clusters initial and terminal point \emph{locations} to produce persistent spatial region polygons. Once estimated during a calibration period, these polygons classify any future trajectory by point-in-polygon containment without re-clustering, reducing the total computational cost from $O(N^2)$ pairwise comparisons to $O(N)$ containment checks. The pipeline proceeds through six steps: preprocessing trajectories to suppress tracking artifacts, estimating a region of interest from observed vehicle activity, selecting initial and terminal trajectory points with optional perspective density equalization, enforcing spatial separation between proximate entry and exit flows via an exclusion zone, clustering the remaining points into coherent groups, and enclosing each cluster within a region polygon. Each step operates without labeled data, and the full pipeline requires no prior knowledge of intersection geometry, camera parameters, or traffic conditions.

The main contributions are:
\begin{itemize}
    \item We address the problem of automated entry and exit region identification for turning movement counts and propose an unsupervised pipeline that delineates region polygons from raw vehicle trajectories, requiring no manual annotation or camera calibration.

    \item We evaluate the pipeline on 25 surveillance cameras deployed across Bengaluru, India, capturing dense heterogeneous traffic, and on 10 sequences from the UA-DETRAC benchmark dataset~\cite{UADETRAC2020}, including validation on 16 held-out camera feeds. We further compare against trajectory clustering approaches from Jana et al.~\cite{Jana2023} and B\'{e}lisle et al.~\cite{Belisle2017}.

    \item We conduct systematic statistical analysis of each pipeline parameter using ANOVA with Tukey HSD, Wilcoxon signed-rank, and Friedman tests with Nemenyi post-hoc comparisons, identifying three parameters that consistently and significantly affect classification accuracy and deriving an empirically grounded recommended configuration.

    \item We conduct an extended evaluation measuring the effects of calibration video duration and traffic density on classification accuracy, providing operational deployment guidance.
\end{itemize}

\textit{To the best of our knowledge, this is the first study to validate fully automated, camera-agnostic entry and exit region identification for vision-based turning movement counts across both heterogeneous traffic conditions and diverse camera deployments.}

The remainder of this manuscript is organized as follows. Section~\ref{sec:related_work} reviews related work. Section~\ref{sec:methodology} details the proposed methodology. Section~\ref{sec:exp_eval} describes the datasets and experimental framework. Section~\ref{sec:results} reports the parametric analysis and overall performance, and Section~\ref{sec:extra_eval} provides extended evaluation under varied operational conditions. Section~\ref{sec:conclusion} concludes with key findings, limitations, and future directions.

%%%%%%%%%%%%%%%%%%%%%%%%%%%%%%%%%%%%%%%%%%%%%%%%%%%%%%%%%%%%%%%%%%%%%%%%%%%%%%%%%%%%%%%%%%%%%%%%%%%%
\section{Related Work}
\label{sec:related_work}

Automation of turning movement counts draws on three interconnected research streams: unsupervised learning of scene structure from trajectory data, applied vision-based TMC pipelines, and data-driven methods for automated zone definition and movement classification. Each stream contributes partial solutions, but none addresses the full problem of automated, camera-agnostic entry and exit region identification. This section reviews each in turn, identifying the limitations that collectively motivate the present work.

\subsection{Unsupervised Scene Structure Learning}

The foundational framework for discovering entry and exit zones from movement patterns was introduced by Makris and Ellis~\cite{Makris2005}, who clustered trajectory endpoints using Gaussian models and demonstrated that semantic scene structure, including paths and stop regions, can be learned without supervision. Wang et al.~\cite{Wang2006} extended this to full semantic scene modeling that identifies entry zones, exit zones, roads, and walkways, notably addressing false entry and exit points arising inside paths by inspecting local density-velocity distributions. Wang et al.~\cite{Wang2011} subsequently introduced Dual Hierarchical Dirichlet Processes that automatically determine both the number of activity categories and the number of semantic regions, eliminating the need to pre-specify zone counts. Morris and Trivedi~\cite{Morris2008} formally defined sources and sinks (entry and exit zones in the scene understanding literature) as key elements of scene topology, and later proposed a three-stage hierarchical framework combining Gaussian Mixture Models, trajectory clustering, and Hidden Markov Models~\cite{Morris2011}.

These works establish that the spatial structure of traffic scenes can, in principle, be recovered from trajectory data alone. However, several limitations constrain their applicability to the deployment context described in Section~\ref{sec:intro}. They require reliable, long-duration trajectory extraction that degrades under heavy occlusion; they assume orderly, lane-following traffic characteristic of the scenes on which they were developed; and they have been demonstrated on at most five camera views without validation across varying camera configurations. None have been operationalized for large-scale TMC deployment.

\subsection{Vision-Based TMC Estimation}

A distinct body of work addresses the applied problem of counting vehicles by turning movement. Shirazi and Morris~\cite{Shirazi2016} combined zone comparison for robust tracking with trajectory comparison, and their survey~\cite{Shirazi2017} remains the definitive reference on vision-based intersection monitoring. More recent work has advanced detection and tracking capabilities while retaining the manual zone-definition requirement. Gloudemans and Work~\cite{Gloudemans2021} presented a localization-based tracking method using manually drawn polygons for classification. Pakdamansavoji et al.~\cite{Pakdamansavoji2024} demonstrated that ground-plane projection improves TMC accuracy, yet their method still requires manual region specification. Entries to the NVIDIA AI City Challenge~\cite{Shuo24AIC24} all operate on pre-defined intersection geometry. Across this line of work, entry and exit zone specification remains a manual, per-intersection step whose cost scales linearly with the number of cameras; it is precisely this requirement that prevents the TMC pipelines described above from being applied at network scale.

\subsection{Automated Zone Definition and Movement Classification}

A smaller body of work has attempted to automate zone definition. Adl et al.~\cite{Adl2024} standardize zones by dividing a fisheye camera view into a central region and peripheral zones. Abdelhalim et al.~\cite{Abdelhalim2020} generate virtual traffic lanes from initial trajectories and classify subsequent ones via $k$-nearest neighbors. These methods, however, rely on assumed intersection geometries or require manual calibration, constraining their applicability in uncalibrated or geometrically complex environments.

Data-driven trajectory clustering has emerged as an alternative that avoids geometric assumptions. Jana et al.~\cite{Jana2023} classify movement types using hierarchical clustering with a custom Hausdorff-based similarity measure, arguing that supervised methods lack the adaptability needed for city-wide deployment. B\'{e}lisle et al.~\cite{Belisle2017} implement a three-step vehicle-counting process based on trajectory clustering using LCSS (Longest Common Subsequence) similarity, representing the closest existing work to fully automated video-based TMC. Their method, however, requires a supervised optimization step in which 15~minutes of manually counted ground-truth TMCs are used to train tracking and grouping parameters, and it requires homography for world-space projection. It was evaluated on three single-lane sites with a weighted average error of approximately 12\%. The proposed pipeline differs from B\'{e}lisle et al.\ in three respects: it is fully unsupervised (no ground-truth counts are needed for calibration), it clusters initial and terminal point \emph{locations} rather than full trajectory shapes (avoiding $O(N^2)$ pairwise comparisons), and it produces persistent spatial region polygons that can classify future trajectories by point-in-polygon containment without re-clustering.

Two further lines of work demonstrate the feasibility of unsupervised spatial learning from traffic data at operational scale, though neither addresses the specific problem of entry and exit region identification for TMC. Qiu et al.~\cite{Qiu2024} deploy an automatic lane learning system for the Indiana Department of Transportation, achieving an F1 score exceeding 0.79 on 45 highway videos. Saldivar-Carranza et al.~\cite{Saldivar-Carranza2021} cluster headings from connected-vehicle GPS data, achieving approximately 98\% accuracy across more than 100 intersections; however, their approach requires GPS data rather than video, limiting its applicability to camera-based surveillance networks.

\subsection{Research Gaps}

Synthesizing across these streams, three gaps emerge from this review that directly motivate the present work. First, no existing method provides fully automated, camera-agnostic entry and exit region identification for TMC. The scene understanding methods~\cite{Makris2005, Wang2006, Wang2011, Morris2011} have not been validated across cameras with varying configurations, and all applied TMC pipelines~\cite{Shirazi2016, Gloudemans2021, Pakdamansavoji2024} require manually specified zones. Second, all validated methods have been evaluated exclusively on lane-following traffic, leaving a documented domain gap~\cite{Mallikarjuna2009} for heterogeneous traffic conditions where vehicles spread laterally across the full road width rather than adhering to discrete lanes. Third, no prior work has conducted systematic statistical evaluation of pipeline parameter sensitivity for this task; existing studies report results under a single configuration without quantifying how design choices affect performance or how robust those choices are across different camera deployments.

%%%%%%%%%%%%%%%%%%%%%%%%%%%%%%%%%%%%%%%%%%%%%%%%%%%%%%%%%%%%%%%%%%%%%%%%%%%%%%%%%%%%%%%%%%%%%%%%%%%%
%%%%%%%%%%%%%%%%%%%%%%%%%%%%%%%%%%%%%%%%%%%%%%%%%%%%%%%%%%%%%%%%%%%%%%%%%%%%%%%%%%%%%%%%%%%%%%%%%%%%
\section{Methodology}
\label{sec:methodology}

\begin{figure*}[!htbp]
    \centering
    \includegraphics[width=\textwidth]{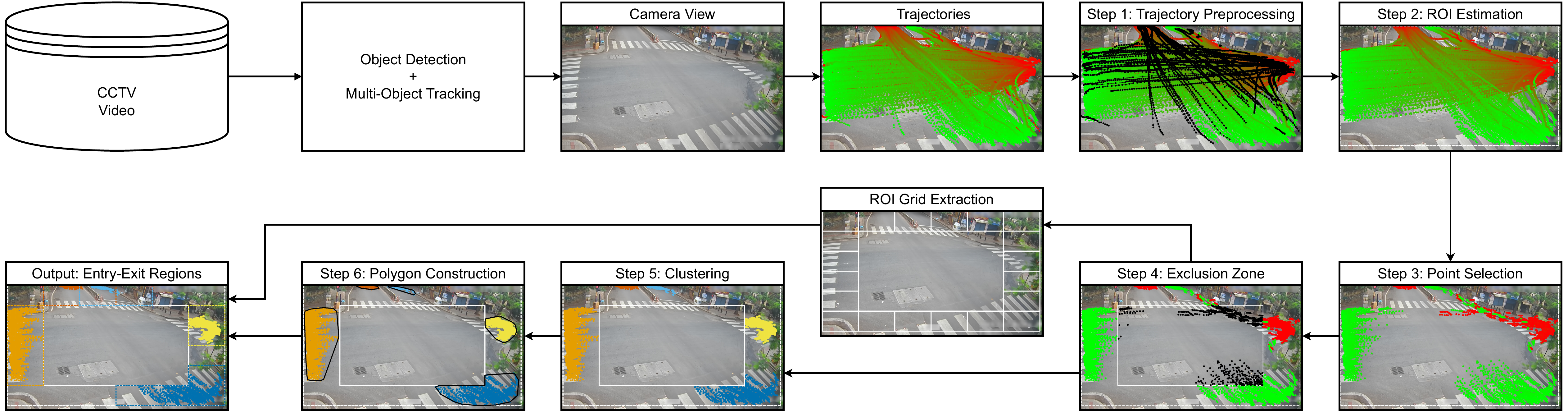}
    \caption{Proposed pipeline for unsupervised entry and exit region identification. The six components progressively refine the raw trajectory data, from filtering tracking artifacts (Step~1) through spatial bounding and point selection (Steps~2--3) to boundary enforcement, clustering, and polygon construction (Steps~4--6). The ROI Grid Extraction block (unnumbered) illustrates the $n \times n$ grid structure derived from the ROI, which is used in grid-based polygon construction (Step~6). The resulting regions are used to classify each trajectory's turning movement.}
    \label{fig:flowchart}
\end{figure*}

The proposed pipeline comprises six sequential components followed by a classification step, each operating without labeled training data or prior knowledge of camera configuration or intersection geometry. The components form a progressive refinement chain: raw trajectories are first cleaned of tracking artifacts, then spatially bounded to the area of observed vehicle activity, reduced to the points most informative for boundary detection, constrained to a boundary zone that enforces separation between proximate entry and exit flows, clustered into coherent spatial groups, and enclosed within polygons that delineate the final entry and exit regions. Several of these components are interdependent; in particular, the boundary zone defined in Step~4 relies on the spatial frame established in Step~2, and the polygon construction in Step~6 is governed by the same boundary zone. Figure~\ref{fig:flowchart} provides an overview of this progression, and Section~\ref{sec:results} evaluates how alternative configurations at each stage affect classification accuracy.

%---------------------------------------------------------------------------------------------------
\subsection*{Step 1: Trajectory Preprocessing}
\label{subsec:preprocessing}

Vehicle tracking in dense, heterogeneous urban traffic is susceptible to errors caused by occlusion. When a larger vehicle temporarily obscures a smaller one, the tracker may assign a new identity upon reappearance, producing a fragmented tracklet whose initial or terminal point falls in the scene interior rather than at the field-of-view boundary. If retained, such fragments bias subsequent clustering toward the intersection interior, because their endpoints do not correspond to genuine entry or exit regions.

To suppress these artifacts, three sequential filters are applied. Let $\mathcal{T}_{raw} = \{T_i\}_{i=1}^{N}$ denote the set of $N$ raw trajectories extracted from the video sequence. Each trajectory $T_i$ is defined by a unique identifier $i$ and an ordered sequence of observations, where each observation comprises a spatial position and a timestamp:
\begin{equation}
    T_i = \bigl((p_1, t_1),\; (p_2, t_2),\; \dots,\; (p_{L_i}, t_{L_i})\bigr), \quad p_j = (x_j, y_j),
\end{equation}
where $p_j$ denotes the bounding-box centroid at the $j$-th observation, $t_j$ is the associated timestamp, and $L_i$ is the trajectory length. Parentheses indicate that the ordering is meaningful. When context makes the trajectory identity clear, the subscript $i$ on $L$ is suppressed.

\textbf{Temporal filtering} discards trajectories shorter than a minimum duration $\tau_{min}$ as likely fragments from identity switches, yielding $\mathcal{T}_{temporal} = \{ T \in \mathcal{T}_{raw} \mid t_L - t_1 \ge \tau_{min} \}$. \textbf{Displacement filtering} removes stationary vehicles by requiring a minimum Euclidean displacement $\delta_{min}$ between initial and terminal positions: $\mathcal{T}_{spatial} = \{ T \in \mathcal{T}_{temporal} \mid \| p_L - p_1 \|_2 \ge \delta_{min} \}$.

\textbf{Neighborhood filtering} removes trajectories whose initial or terminal points are spatially isolated, as these are unlikely to represent genuine entry or exit flows. Let $\mathcal{X}$ denote the global pool of all initial and terminal points across $\mathcal{T}_{spatial}$, and let $\eta(q, r) = | \{ x \in \mathcal{X} \mid \| q - x \|_2 \le r \} |$ count the number of pool members within radius $r$ of a query point $q$, including $q$ itself. A trajectory is retained only if both its initial and terminal points exceed the minimum neighbor count $n_{min}$:
\begin{equation}
    \mathcal{T}_{filtered} = \left\{ T_i \in \mathcal{T}_{spatial} \;\middle|\;
    \begin{aligned}
    &\eta(p_1^{(i)}, r) > n_{min} \\
    \land\;\; &\eta(p_L^{(i)}, r) > n_{min}
    \end{aligned}
    \right\}.
\end{equation}

The values of $\tau_{min}$, $\delta_{min}$, $r$, and $n_{min}$ are fixed across all deployments and are specified, with empirical justification, in the Supplementary Material (Section~S-I).

%---------------------------------------------------------------------------------------------------
\subsection*{Step 2: Region of Interest Estimation}
\label{subsec:ROI}

Traffic cameras in uncontrolled deployments are frequently mounted on the side of the road rather than directly above the carriageway. As a result, a vehicle entering the field of view from the far side may first appear well inside the image frame rather than at its edge (cf.\ Fig.~S5 in the Supplementary Material). If entry and exit regions are defined relative to the full image boundary, such vehicles would appear to originate from the scene interior, confounding subsequent analysis. More broadly, the camera's field of view typically encompasses areas where vehicles cannot travel, including sidewalks, vegetation, and signage. Including these areas would enlarge the spatial search domain and, critically, would misalign the exclusion zone defined in Step~4: because that zone is centered on the region of interest, an incorrectly bounded region propagates errors into the separation of entry and exit clusters.

To restrict analysis to the area of observed vehicle activity, the bounding rectangle of all trajectory coordinates in $\mathcal{T}_{filtered}$ is computed:
\begin{equation}
    \mathcal{R} = [x_{min},\; x_{max}] \times [y_{min},\; y_{max}],
\end{equation}
where the bounds are the coordinate extrema over all points in the filtered trajectory set. This rectangle defines the region of interest~(ROI) for all subsequent steps, with width $W_{\mathcal{R}}$ and height $H_{\mathcal{R}}$. A rectangular shape is chosen deliberately: the exclusion zone in Step~4 is also rectangular and is defined as a concentric sub-rectangle of~$\mathcal{R}$, so a consistent geometry simplifies their coupling. From this point forward, ``boundary'' refers to the periphery of~$\mathcal{R}$ rather than the edges of the full image frame.

%---------------------------------------------------------------------------------------------------
\subsection*{Step 3: Initial and Terminal Point Selection}
\label{subsec:no_of_points}

The initial and terminal points of a trajectory carry the most direct information about where a vehicle entered and exited the field of view. Interior points, by contrast, are distributed across the road surface and primarily encode path information rather than boundary location. Including all trajectory points in the clustering step introduces two problems. First, the spatial dispersion of interior points dilutes the concentration of points at the ROI boundary, obscuring the entry and exit signal. Second, in cameras mounted at low height or steep pitch, the projected distance between adjacent entry and exit regions can be small; additional interior points inflate each cluster's spatial extent, making neighboring regions harder to separate.

To isolate boundary information, a selection function $\Psi(T, s)$ extracts the first and last $s$ spatial positions from each trajectory $T = (p_1, \dots, p_L)$ (timestamps are not used beyond Step~1), where $s$ is the number of points selected per end (the formal definition is provided in Supplementary Material, Section~S-IV). When $s = 1$, the function returns only the single initial and single terminal point; when $s \ge \lfloor L/2 \rfloor$, the two subsequences overlap or exhaust the trajectory, and all points are retained. The selected points are aggregated across all filtered trajectories:
\begin{equation}
    \mathcal{P}_{selected} = \bigcup_{T \in \mathcal{T}_{filtered}} \Psi(T, s).
\end{equation}
The values of $s$ evaluated in this study are specified in Section~\ref{sec:exp_eval}.

Because traffic cameras are tilted downward toward the road surface, initial and terminal points originating from the far field are packed more densely in pixel space than those from the near field. If left unaddressed, this density gradient could bias clustering toward over-segmenting dense far-field regions while under-segmenting sparser near-field regions. To investigate whether correcting this imbalance improves region detection, an optional perspective density equalization is evaluated as a binary parameter. When enabled, the procedure partitions the vertical extent of the ROI into horizontal bands whose widths are determined by the Freedman--Diaconis rule, and reduces the point count in high-density bands via stratified sampling. Its effect on classification accuracy is reported in Section~\ref{sec:results}.

%---------------------------------------------------------------------------------------------------
\subsection*{Step 4: Exclusion Zone}
\label{subsec:buffer_size}

When a camera is mounted at low height, at steep pitch, or with a narrow field of view, the projected distance between an entry region on one side of the intersection and the corresponding exit region on the opposite side can be small in the image plane. In such configurations, initial and terminal points from distinct regions may overlap spatially, preventing clustering algorithms from separating them into distinct groups. The primary purpose of the exclusion zone is to enforce spatial separation by removing the central portion of the ROI where this intermingling occurs, thereby restricting the point set to a peripheral buffer where entry and exit flows are spatially distinguishable.

Let the exclusion zone $\mathcal{R}_{excl}$ be a rectangle concentric with the ROI $\mathcal{R}$, with margins $m_x$ and $m_y$ inset along the horizontal and vertical axes respectively:
\begin{equation}
\begin{split}
    \mathcal{R}_{excl} = \; &[x_{min} + m_x,\; x_{max} - m_x] \\
    &\times [y_{min} + m_y,\; y_{max} - m_y].
\end{split}
\end{equation}
The margins are expressed as fractions of the ROI dimensions. For a given exclusion ratio $1/n$, where $n$ is a positive integer, the margins are $m_x = W_{\mathcal{R}} / n$ and $m_y = H_{\mathcal{R}} / n$, where $W_{\mathcal{R}}$ and $H_{\mathcal{R}}$ denote the width and height of the ROI. The buffer region $\mathcal{B} = \mathcal{R} \setminus \mathcal{R}_{excl}$ retains only points in the peripheral zone:
\begin{equation}
    \mathcal{P}_{boundary} = \bigl\{ p \in \mathcal{P}_{selected} \mid p \notin \mathcal{R}_{excl} \bigr\}.
\end{equation}

The exclusion ratio also governs the grid resolution used in Step~6: a ratio of $1/n$ induces an $n \times n$ grid over the ROI. Because each margin equals exactly one cell width, the exclusion zone boundary aligns with the grid lines: the interior $(n{-}2) \times (n{-}2)$ cells fall within the exclusion zone, and the outermost ring of cells, one cell deep on each side, forms the buffer zone $\mathcal{B}$. For example, when $n = 6$, the ROI is divided into 36 cells; the inner $4 \times 4 = 16$ cells form the exclusion zone, and the remaining 20 peripheral cells form the buffer. Increasing $n$ therefore simultaneously narrows the buffer (more aggressive center removal) and refines the grid (more, smaller cells). These two parameters are coupled rather than independent, and this coupling is evaluated as a single design choice. The specific ratios tested are listed in Section~\ref{sec:exp_eval}.

Two aspects of this step merit emphasis. First, its effectiveness depends on the ROI computed in Step~2. If the ROI is defined relative to the full image rather than the area of observed vehicle activity, the exclusion zone may not coincide with the actual intersection interior, and the intended separation effect is lost. Second, the exclusion zone is not discarded after this step; it persists as a geometric constraint in Step~6, where it determines hull clipping boundaries and grid cell candidacy.

%---------------------------------------------------------------------------------------------------
\subsection*{Step 5: Clustering}
\label{subsec:clustering}

The points in $\mathcal{P}_{boundary}$ must be grouped into coherent clusters corresponding to distinct entry and exit regions. From this step onward, only the spatial coordinates $(x, y)$ of each point are used; the temporal component $t_j$, which served the filtering criteria in Step~1, plays no further role.

Clustering is performed separately on the initial points and terminal points within $\mathcal{P}_{boundary}$. Initial points identify candidate entry regions; terminal points identify candidate exit regions. Separate clustering prevents entry and exit regions with overlapping spatial footprints from being merged, particularly in heterogeneous traffic where they may coincide when projected onto the image plane.

Three clustering algorithms are evaluated, namely K-Means, Gaussian Mixture Models (GMM), and DBSCAN, each making different assumptions about cluster shape and density structure.

\emph{K-Means}~\cite{MacQueen1967} partitions points into $K$ clusters by minimizing within-cluster variance, producing compact, roughly spherical cluster shapes. \emph{DBSCAN}~\cite{Ester1996} defines clusters on the basis of local point density; it naturally labels low-density points as noise, which is advantageous when residual artifacts survive preprocessing. \emph{GMM}~\cite{McLachlan2000} models the data as a mixture of $K$ Gaussian components with full covariance matrices, fitted via Expectation-Maximization~(EM)~\cite{Dempster1977}. The bounds of the cluster search spaces and automatic parameter selection procedures are specified in Section~\ref{sec:exp_eval}.

%---------------------------------------------------------------------------------------------------
\subsection*{Step 6: Region Polygon Construction}
\label{subsec:grid_or_hull}

Each cluster identified in Step~5 must be enclosed by a polygon to delineate the corresponding entry or exit region. Two representations are evaluated: convex hulls and grid-based polygons.

\textbf{Convex hulls.} The convex hull~\cite{Barber1996} of each cluster $C_j$ is computed as the smallest convex polygon $H_j$ containing all points assigned to that cluster. Hulls provide tight, data-driven boundaries but are limited to the spatial footprint of the trajectories observed during the calibration period. If vehicles during this period do not cover the full lane width, the resulting hull will be too narrow and will fail to capture future vehicles that travel along paths not represented in the calibration data. Even if the central exclusion zone contains no points, a calculated convex hull might still bridge across it; in such cases, the hull is clipped to the buffer zone: $H_j' = H_j \setminus \mathcal{R}_{excl}$.

\textbf{Grid-based polygons.} To address the coverage limitation of convex hulls, a grid-based alternative is evaluated. As noted in Step~4, the exclusion ratio $1/n$ determines an $n \times n$ grid over the ROI, with each cell having dimensions $W_{\mathcal{R}}/n \times H_{\mathcal{R}}/n$. Only cells that lie within the buffer zone $\mathcal{B}$ are candidates for assignment. Each candidate cell $g_i$ is assigned to the hull $H_j$ with which it shares the greatest number of data points:
\begin{equation}
    \phi(g_i) = H_{j^*}, \quad \text{where} \quad
    j^* = \operatorname*{argmax}_{j} \bigl| \mathcal{P}_{boundary} \cap g_i \cap H_j \bigr|.
\end{equation}
Cells assigned to the same hull are merged into a single region polygon: $R_j = \bigcup_{\phi(g_i) = H_j} g_i$. Because grid cells are uniformly sized, the resulting regions extend beyond the convex hull to cover areas where vehicles may plausibly travel but were not observed during the calibration period.

It is important to note two properties of the grid-based approach here. First, the grid resolution is coupled to the exclusion ratio: both are governed by the single parameter $n$. Second, the grid relies on the exclusion zone to constrain regions to the ROI periphery. Without an exclusion zone, grid cells near the image center may be assigned to entry or exit regions, producing regions that extend into the intersection interior rather than remaining at the boundary.

%---------------------------------------------------------------------------------------------------
\subsection*{Turning Movement Classification}
\label{subsec:classification}

Once entry and exit regions have been delineated, the preprocessed trajectories from $\mathcal{T}_{filtered}$ are classified by assigning each trajectory's initial point to the entry region in which it falls and its terminal point to the corresponding exit region. The resulting entry--exit pair defines the turning movement for that trajectory. Trajectories whose initial or terminal points do not fall within any identified region remain unclassified. This classification step requires no additional parameters; it depends entirely on the region polygons produced by the preceding steps.

%%%%%%%%%%%%%%%%%%%%%%%%%%%%%%%%%%%%%%%%%%%%%%%%%%%%%%%%%%%%%%%%%%%%%%%%%%%%%%%%%%%%%%%%%%%%%%%%%%%%
%%%%%%%%%%%%%%%%%%%%%%%%%%%%%%%%%%%%%%%%%%%%%%%%%%%%%%%%%%%%%%%%%%%%%%%%%%%%%%%%%%%%%%%%%%%%%%%%%%%%
\section{Experimental Setup}
\label{sec:exp_eval}

The methodology described in Section~\ref{sec:methodology} introduces configurable choices at five of its six steps; only Step~1 (trajectory preprocessing) uses fixed parameters. To evaluate these choices systematically, the pipeline is executed under 900 configurations across 19 camera feeds spanning two datasets and two distinct traffic regimes. This section describes the datasets (Section~\ref{subsec:dataset}), the experimental grid that defines the parameter space (Section~\ref{subsec:exp_grid}), the evaluation metric that measures classification accuracy against manually annotated ground truth (Section~\ref{subsec:eval_metric}), and the statistical testing framework that determines which parameter differences are significant after accounting for variability across camera feeds (Section~\ref{subsec:stat_tests}).

\subsection{Datasets}
\label{subsec:dataset}

The pipeline is evaluated on two complementary data sources spanning different traffic conditions, camera configurations, and annotation protocols.

\textbf{Bengaluru Safe City Camera Dataset.} The primary dataset comprises video from 25 surveillance cameras selected from a pool of approximately 5{,}000 Safe City cameras deployed across Bengaluru, India. Cameras were selected on the basis of field-of-view coverage: each selected camera encompasses an intersection or high-volume road segment. Since these cameras serve general public safety, their mounting angles, fields of view, and positioning are not optimized for traffic engineering. All videos were captured at $1920 \times 1080$ resolution at 25--50~FPS.

Of the 25 cameras, nine were designated for pipeline development and parametric analysis, with video available from three different days at each location, yielding 11 to 13 clips of 15~minutes each per day. The parametric analysis in Section~\ref{sec:results} uses a single randomly selected 15-minute clip per camera, all recorded on the same day to ensure comparability across locations. The extended evaluation in Section~\ref{sec:extra_eval} estimates regions from one day's data and evaluates on the remaining days, with clips selected at random. Separately, video from an additional day outside the three development days was collected for all 25 cameras (the 9 development locations and 16 held-out locations). This temporally independent video is used for the overall performance assessment (Section~\ref{sec:results}) and the trajectory clustering comparison, ensuring that even the development cameras are evaluated on data not seen during parameter selection.

Object detection was performed using RT-DETR~\cite{RTDETR2024}, trained on the Indian Driving Dataset~\cite{IDD2019} and the Bengaluru Mobility Challenge~\cite{BMC2024}. Multi-object tracking used BoT-SORT-ReID~\cite{BoTSORT2022}. The reported classification errors measure the pipeline's ability to assign detected-and-tracked vehicles to the correct turning movement; upstream detection and tracking errors are not disaggregated. Ground-truth entry and exit polygons were annotated by three domain experts, with disagreements resolved through joint review. This dataset is planned for public release as a separate publication, providing one of the first vehicle trajectory datasets from high-density, heterogeneous Indian traffic with TMC annotations.

\textbf{UA-DETRAC Benchmark Dataset.} To evaluate generalizability beyond the conditions represented by the Bengaluru data, 10 sequences were selected from the UA-DETRAC benchmark dataset~\cite{UADETRAC2020}, comprising over 140{,}000 frames captured at 24 locations in Beijing and Tianjin, China. Entry and exit regions were manually annotated following the same protocol. Since UA-DETRAC provides manually verified tracking annotations, the preprocessing step (Step~1) was not applied to these sequences; raw trajectories enter Step~2 directly. This asymmetry is noted when interpreting cross-dataset comparisons.

\textbf{Cross-dataset considerations.} The Bengaluru data features uncalibrated cameras with oblique mounting angles, dense heterogeneous traffic with frequent occlusion, and automatically generated trajectory data that includes noise from identity switches. The UA-DETRAC data features relatively consistent camera perspectives, lane-following traffic, and cleaner manually verified annotations. These contrasting conditions allow the parametric analysis to reveal which pipeline components are sensitive to data quality and traffic regime. Camera view diversity across the 25 Bengaluru locations is substantiated in the Supplementary Material (Section~S-VIII) using DINOv3 embeddings.

\subsection{Experimental Grid}
\label{subsec:exp_grid}

Each configurable choice within the pipeline is treated as an experimental variable with a discrete set of evaluated levels. The complete set of variables and their levels is summarized in Table~\ref{tab:experiment_grid}. Two parameters are coupled rather than independent: as described in Step~4, the exclusion ratio $1/n$ simultaneously determines the exclusion zone margins and the $n \times n$ grid resolution used in Step~6; they are therefore evaluated as a single design choice. The preprocessing parameters from Step~1 ($\tau_{min}$, $\delta_{min}$, $r$, $n_{min}$) were held fixed across all configurations; their values and empirical justification are provided in the Supplementary Material (Section~S-I).

The pipeline was executed under all valid combinations of the six experimental variables for every camera feed in both datasets, yielding $2 \times 6 \times 2 \times 7 \times 3 \times 2 = 1{,}008$ total combinations per feed. Of these, 108 were excluded: 72 invalid combinations (grid-based polygon construction requires the exclusion ratio parameter $n$, which is undefined when no exclusion zone is applied) and 72 repeated combinations (without an exclusion zone, ROI estimation has no effect on the remaining pipeline steps), with 36 falling in both categories. This yields $1{,}008 - 72 - 72 + 36 = 900$ valid unique configurations per feed. With 9 Bengaluru development cameras and 10 UA-DETRAC sequences, this produces 17{,}100 total pipeline executions.

\begin{table}[!htbp]
    \centering
    \caption{Experimental grid evaluated across all camera feeds. Each row corresponds to a configurable pipeline step; all valid combinations were tested. The exclusion ratio and grid resolution are coupled (Step~4).}
    \label{tab:experiment_grid}
    \resizebox{\columnwidth}{!}{\begin{tabular}{@{}lll@{}}
        \toprule
        \textbf{Pipeline Step} & \textbf{Parameter} & \textbf{Evaluated Levels} \\
        \midrule
        Step 2: ROI Estimation
            & ROI applied
            & Yes, No \\
        Step 3: Point Selection
            & Points per end ($s$)
            & 1, 3, 5, 7, 9, All (per-trajectory) \\
        Step 3: Density Equalization
            & Subsampling applied
            & Yes, No \\
        Step 4: Exclusion Zone
            & Exclusion ratio ($1/n$)
            & None, 1/3, 1/4, 1/5, 1/6, 1/7, 1/8 \\
        Step 5: Clustering
            & Algorithm
            & K-Means, DBSCAN, GMM\textsuperscript{*} \\
        Step 6: Polygon Construction
            & Representation
            & Grid, Convex hull \\
        \bottomrule
    \end{tabular}}
    \raggedright
    \footnotesize
    \textsuperscript{*}Clustering parameters are determined automatically: $K$ for K-Means and $\epsilon$ for DBSCAN (with $MinPts = 2D = 4$ for $D = 2$ dimensions) are found via knee-point detection~\cite{Satopaa2011} on their sum-of-squares and $MinPts$-NN distance curves, respectively. The optimal number of GMM components is selected by minimizing the BIC~\cite{Schwarz1978}. The search space for both $K$ and the number of GMM components is bounded between $2$ and $9$.
\end{table}

\subsection{Evaluation Metric}
\label{subsec:eval_metric}

The pipeline's objective is accurate classification of trajectories by turning movement, not geometric fidelity of the predicted regions. Metrics commonly used for region-based evaluation in computer vision, such as Intersection over Union (IoU) and mean average precision (mAP)~\cite{Zou2019, Minaee2022}, evaluate geometry alone and do not capture whether trajectories are correctly assigned to their respective movements. A geometrically smaller but functionally correct region could receive a low IoU score if the ground-truth annotation includes margins beyond the visible road surface. The evaluation therefore uses a task-specific trajectory assignment error that directly measures the fraction of trajectories misclassified for each turning movement.

\textbf{Polygon label association.} Because the predicted polygons carry no inherent labels, a mapping procedure associates each predicted polygon with a ground-truth polygon on the basis of shared trajectory points. Let $\mathcal{P}_{pred} = \{P_1, \dots, P_m\}$ denote predicted polygons and $\mathcal{G} = \{G_1, \dots, G_q\}$ the ground-truth polygons. For each predicted polygon $P_i$, the assigned label is that of the ground-truth polygon sharing the greatest number of initial or terminal trajectory points:
\begin{equation}
\begin{split}
    L(P_i) &= \text{label}(G_{j^*}), \\
    \text{where} \quad j^* &= \operatorname*{argmax}_{j \in \{1, \dots, q\}}
    \bigl| \mathcal{P}_{boundary} \cap P_i \cap G_j \bigr|,
\end{split}
\end{equation}
where $\mathcal{P}_{boundary}$ is the set of boundary points defined in Step~4. The mapping is many-to-one: multiple predicted polygons may receive the same ground-truth label, which occurs when clustering over-segments a single entry or exit region.

\textbf{Trajectory assignment error.} Each preprocessed trajectory from $\mathcal{T}_{filtered}$ is assigned an entry--exit label pair: the entry label is determined by the region polygon containing the trajectory's initial point, and the exit label by the polygon containing its terminal point. The resulting pair defines the trajectory's predicted turning movement, and ground-truth movements are determined analogously using the ground-truth polygons.

A turning movement is defined as an origin-destination pair $(o, d)$, where $o$ indicates the entry region and $d$ the exit region. For each turning movement $(o, d)$, trajectory assignment is evaluated using a one-versus-rest formulation. A trajectory is a true positive ($TP$) for movement $(o, d)$ if both the predicted and ground-truth labels assign it to $(o, d)$; a false positive ($FP$) if the predicted label assigns it to $(o, d)$ but the ground-truth label does not; and a false negative ($FN$) if the ground-truth label assigns it to $(o, d)$ but the predicted label does not. The per-movement error is:
\begin{equation}
    E_{(o,d)} = \frac{FP_{(o,d)} + FN_{(o,d)}}{TP_{(o,d)} + FP_{(o,d)} + FN_{(o,d)} + TN_{(o,d)}}.
\end{equation}
The denominator equals the total number of trajectories in the evaluation set. Let $\mathcal{M}$ denote the set of all valid origin-destination pairs at the intersection, with $M = |\mathcal{M}|$. The overall classification error for a given camera feed is the macro-average across all $M$ turning movements: $\bar{E} = \frac{1}{M} \sum_{(o,d) \in \mathcal{M}} E_{(o,d)}$.

\subsection{Statistical Testing Framework}
\label{subsec:stat_tests}

The experimental grid produces 900 configurations per camera feed, and raw error values vary across feeds due to differences in camera geometry, traffic density, and trajectory quality. Comparing configurations by their mean error alone does not account for this variability. Statistical testing is therefore used to determine which parameter differences are significant after accounting for feed-level variation.

Two complementary families of tests are employed. Parametric analysis uses one-way ANOVA~\cite{Cuevas2004} with Tukey HSD~\cite{Abdi2010} post-hoc comparisons at $\alpha = 0.05$. Each pipeline parameter is tested independently; interactions between parameters are not modeled, a simplification discussed further in Section~\ref{sec:conclusion}. However, the trajectory assignment error $E_{(o,d)}$ is bounded in $[0, 1]$ and tends to be right-skewed~\cite{Limpert2011}, potentially violating the normality and homoscedasticity assumptions underlying ANOVA. Nonparametric tests are therefore conducted in parallel: the Wilcoxon signed-rank test~\cite{Taheri2013} for two-level parameters and the Friedman test~\cite{Pereira2015} with Nemenyi post-hoc comparisons~\cite{Nemenyi1963} for multi-level parameters. Where the two frameworks agree, confidence in the finding is strengthened; where they diverge, the nonparametric result is given precedence as the more conservative estimate.

%%%%%%%%%%%%%%%%%%%%%%%%%%%%%%%%%%%%%%%%%%%%%%%%%%%%%%%%%%%%%%%%%%%%%%%%%%%%%%%%%%%%%%%%%%%%%%%%%%%%
%%%%%%%%%%%%%%%%%%%%%%%%%%%%%%%%%%%%%%%%%%%%%%%%%%%%%%%%%%%%%%%%%%%%%%%%%%%%%%%%%%%%%%%%%%%%%%%%%%%%
\section{Results}
\label{sec:results}

Rather than presenting individual error values for every configuration in the experimental grid, the statistical testing framework described in Section~\ref{subsec:stat_tests} is applied to the 17{,}100 pipeline executions (900 configurations $\times$ 19 camera feeds; see Section~\ref{subsec:exp_grid}) to identify which parameters most strongly influence classification error and to determine an empirically grounded recommended configuration. The parametric and nonparametric analyses are presented in turn, followed by a synthesis of the findings, the overall performance under the recommended configuration, and a comparison with trajectory clustering baselines.

\subsection{Parametric Analysis: ANOVA with Tukey HSD}

Tables~\ref{tab:anova_tukey_combined_aicoe} and~\ref{tab:anova_tukey_combined_uad} present the one-way ANOVA results and Tukey HSD post-hoc comparisons for each parameter on both datasets. In the Tukey HSD column, ``$<$'' indicates that the left configuration achieved significantly lower classification error at $\alpha = 0.05$. Parameters are listed in pipeline step order; statistically significant $p$-values ($\alpha = 0.05$) are shown in bold. The complete pairwise comparison matrices are provided in the Supplementary Material (Sections~S-V and~S-VI).

\begin{table}[!htbp]
\centering
\caption{One-way ANOVA and Tukey HSD post-hoc analysis on the Bengaluru dataset. Only statistically significant pairwise differences ($\alpha = 0.05$) are reported. Step numbers refer to Section~\ref{sec:methodology}.}
\label{tab:anova_tukey_combined_aicoe}
\resizebox{\columnwidth}{!}{
\begin{tabular}{@{}lll@{}}
\toprule
\textbf{Parameter (Step)} & \textbf{ANOVA ($p$-value)} & \textbf{Tukey HSD} \\
\midrule
ROI estimation (2) & $7.609 \times 10^{-1}$ &
    No significant pairwise differences \\
Point selection (3) & $\mathbf{7.128 \times 10^{-82}}$ &
    $s \in \{1,3,5,7,9\}$ $<$ All \\
Density equalization (3) & $3.007 \times 10^{-1}$ &
    No significant pairwise differences \\
Exclusion zone (4) & $\mathbf{4.881 \times 10^{-29}}$ &
    \{None \dots 1/7\} $<$ 1/8 \\
Clustering algorithm (5) & $\mathbf{7.876 \times 10^{-21}}$ &
    K-Means $<$ GMM $<$ DBSCAN \\
Polygon representation (6) & $7.161 \times 10^{-2}$ &
    No significant pairwise differences \\
\bottomrule
\end{tabular}}
\end{table}

\begin{table}[!htbp]
\centering
\caption{One-way ANOVA and Tukey HSD post-hoc analysis on the UA-DETRAC dataset. Only statistically significant pairwise differences ($\alpha = 0.05$) are reported.}
\label{tab:anova_tukey_combined_uad}
\resizebox{\columnwidth}{!}{
\begin{tabular}{@{}lll@{}}
\toprule
\textbf{Parameter (Step)} & \textbf{ANOVA ($p$-value)} & \textbf{Tukey HSD} \\
\midrule
ROI estimation (2) & $\mathbf{3.130 \times 10^{-57}}$ &
    With ROI $<$ without ROI \\
Point selection (3) & $\mathbf{3.300 \times 10^{-35}}$ &
    $s \in \{7,9\}$ $<$ $s \in \{3, \text{All}\}$ $<$ $s = 1$; $s{=}5$ n.s.\ from either group \\
Density equalization (3) & $\mathbf{4.432 \times 10^{-18}}$ &
    Without subsampling $<$ with subsampling \\
Exclusion zone (4) & $\mathbf{1.234 \times 10^{-100}}$ &
    \{None, 1/3, 1/4, 1/5\} $<$ 1/6 $<$ 1/7 $<$ 1/8 \\
Clustering algorithm (5) & $\mathbf{9.728 \times 10^{-21}}$ &
    K-Means $<$ GMM $<$ DBSCAN \\
Polygon representation (6) & $\mathbf{1.762 \times 10^{-4}}$ &
    Grid $<$ convex hull \\
\bottomrule
\end{tabular}}
\end{table}

Three parameters achieve high significance on the Bengaluru dataset (Table~\ref{tab:anova_tukey_combined_aicoe}), all with $p < 0.001$. Selecting any number of initial and terminal points per end ($s \in \{1, 3, 5, 7, 9\}$) significantly outperforms using all trajectory points, with no significant pairwise differences among the restricted selections. This indicates that even $s = 1$ suffices to capture the entry and exit signal on this dataset. The exclusion zone exhibits a threshold effect: classification error increases significantly only at the narrowest buffer configuration ($1/n = 1/8$), where the buffer zone is too thin to retain all genuine boundary points; all wider buffer configurations, including no exclusion zone, are statistically indistinguishable. K-Means consistently achieves the lowest error, followed by GMM and then DBSCAN. The remaining three parameters, ROI estimation, density equalization, and polygon representation, do not produce significant differences, suggesting robustness to these choices on this dataset.

On the UA-DETRAC dataset (Table~\ref{tab:anova_tukey_combined_uad}), all six parameters achieve significance at $p < 0.001$, with a finer-grained structure. Selecting 7 or 9 points per end yields the lowest error, significantly outperforming both $s = 3$ and using all points; $s = 5$ is not significantly different from either group and bridges the two. All restricted selections significantly outperform $s = 1$. The K-Means $<$ GMM $<$ DBSCAN hierarchy is consistent with the Bengaluru findings. Configurations with wider buffer zones, including no exclusion zone, significantly outperform those with narrower buffers ($n \ge 6$). In contrast to the Bengaluru results, ROI estimation significantly reduces error, subsampling increases error, and the grid-based representation outperforms convex hulls.

The divergence in significance across datasets is consistent with differences in their data characteristics. The UA-DETRAC cameras frequently frame scenes with large non-road areas, which makes ROI estimation consequential for correctly positioning the exclusion zone. The cleaner trajectory data in UA-DETRAC means that subsampling discards informative points without compensating for meaningful density imbalance. Well-defined cluster boundaries in this dataset allow the standardized geometry of grid cells to provide a measurable advantage over data-dependent convex hulls. On the Bengaluru dataset, the noisier, spatially diffuse trajectory data from heterogeneous traffic may mask these subtler effects, though this interpretation remains a hypothesis rather than a measured outcome.

\subsection{Nonparametric Analysis: Friedman and Wilcoxon Tests}

Tables~\ref{tab:friedman_aicoe} and~\ref{tab:friedman_uad} present the nonparametric results, which verify robustness of the parametric findings under the bounded, right-skewed error distributions characteristic of this task.

\begin{table*}[!htbp]
\centering
\caption{Effect of pipeline parameters on classification error (ranked) on the Bengaluru dataset.}
\label{tab:friedman_aicoe}
\begin{tabular}{@{}llcllc@{}}
\toprule
\textbf{Parameter (Step)} & \textbf{Evaluated Levels (Ranked)} & \textbf{Mean Rank} &
\textbf{Statistical Test} & \textbf{Test Statistic} &
\textbf{\textit{p}-value} \\
\midrule
\textbf{ROI estimation (2)} & With ROI, Without ROI &
    1.44, 1.56 &
    Wilcoxon & $V = 17.00$ & 0.570 \\
\textbf{Point selection (3)} & $s$: 7, 9, 3, 1, 5, All &
    2.44, 2.78, 3.00, 3.33, 3.44, 6.00 &
    Friedman & $\chi^2(5) = 21.00$ & \textbf{0.001} \\
\textbf{Density equal.\ (3)} & With, Without subsampling &
    1.44, 1.56 &
    Wilcoxon & $V = 22.00$ & 1.000 \\
\textbf{Exclusion zone (4)} & $1/n$: 1/6, 1/5, 1/4, 1/7, 1/3, 1/8, None &
    2.67, 2.89, 3.44, 3.78, 4.89, 5.00, 5.33 &
    Friedman & $\chi^2(6) = 13.38$ & \textbf{0.037} \\
\textbf{Clustering algorithm (5)} & K-Means, GMM, DBSCAN &
    1.56, 1.67, 2.78 &
    Friedman & $\chi^2(2) = 8.22$ & \textbf{0.016} \\
\textbf{Polygon repr.\ (6)} & Grid, Convex hull &
    1.33, 1.67 &
    Wilcoxon & $V = 12.00$ & 0.250 \\
\bottomrule
\end{tabular}
\end{table*}

\begin{table*}[!htbp]
\centering
\caption{Effect of pipeline parameters on classification error (ranked) on the UA-DETRAC dataset.}
\label{tab:friedman_uad}
\begin{tabular}{@{}llcllc@{}}
\toprule
\textbf{Parameter (Step)} & \textbf{Evaluated Levels (Ranked)} & \textbf{Mean Rank} &
\textbf{Statistical Test} & \textbf{Test Statistic} &
\textbf{\textit{p}-value} \\
\midrule
\textbf{ROI estimation (2)} & With ROI, Without ROI &
    1.20, 1.80 &
    Wilcoxon & $V = 10.00$ & 0.084 \\
\textbf{Point selection (3)} & $s$: 9, 7, 5, 3, All, 1 &
    2.20, 2.50, 3.50, 3.90, 4.40, 4.50 &
    Friedman & $\chi^2(5) = 13.31$ & \textbf{0.021} \\
\textbf{Density equal.\ (3)} & Without, With subsampling &
    1.00, 2.00 &
    Wilcoxon & $V = 0.00$ & \textbf{0.002} \\
\textbf{Exclusion zone (4)} & $1/n$: 1/4, 1/3, 1/5, None, 1/6, 1/7, 1/8 &
    2.50, 3.10, 3.10, 3.50, 3.90, 5.50, 6.30 &
    Friedman & $\chi^2(6) = 25.36$ & \textbf{$<$0.001} \\
\textbf{Clustering algorithm (5)} & K-Means, GMM, DBSCAN &
    1.40, 2.00, 2.50 &
    Friedman & $\chi^2(2) = 6.21$ & \textbf{0.045} \\
\textbf{Polygon repr.\ (6)} & Grid, Convex hull &
    1.20, 1.80 &
    Wilcoxon & $V = 13.00$ & 0.160 \\
\bottomrule
\end{tabular}
\end{table*}

On the Bengaluru dataset (Table~\ref{tab:friedman_aicoe}), the Nemenyi post-hoc test confirms that using all trajectory points (All) produces significantly higher error ranks than all restricted selections ($p < 0.05$), with no significant differences among the restricted selections. Among the restricted selections, $s = 7$ achieves the lowest mean rank (2.44). Both K-Means (Nemenyi $p = 0.026$) and GMM (Nemenyi $p = 0.048$) achieve significantly lower error ranks than DBSCAN; however, the K-Means versus GMM comparison does not reach Nemenyi significance, a partial divergence from the parametric result where all three pairs were significant. The exclusion zone shows a significant omnibus effect, with ratios near $1/6$ achieving the lowest mean rank, though no specific pairwise comparison exceeds the Nemenyi significance threshold. The remaining parameters confirm the pattern observed in the parametric analysis: no significant effect for polygon representation, ROI estimation, or density equalization.

On UA-DETRAC (Table~\ref{tab:friedman_uad}), density equalization produces the strongest effect ($V = 0.00$, $p = 0.002$): omitting subsampling is uniformly superior. The Nemenyi test confirms that the exclusion ratio $1/4$ significantly outperforms both $1/7$ (Nemenyi $p = 0.031$) and $1/8$ (Nemenyi $p = 0.001$), consistent with the parametric finding that wider buffer zones outperform narrower ones. K-Means achieves significantly lower error ranks than DBSCAN (Nemenyi $p = 0.037$). ROI estimation approaches but does not reach conventional significance ($p = 0.084$), a partial divergence from the parametric result where it was highly significant; the Wilcoxon test's lower statistical power with the smaller effective sample size may explain this difference.

The cross-dataset pattern in the nonparametric analysis confirms the parametric findings: the same three parameters (point selection, clustering algorithm, and exclusion zone) achieve significance on both datasets, while ROI estimation, density equalization, and polygon representation achieve significance only on UA-DETRAC in the parametric analysis, with ROI estimation and polygon representation falling short of significance even in the nonparametric analysis. This pattern is consistent with the interpretation that these parameters have genuine but moderate effects that are detectable only in cleaner data and with more powerful statistical tests.

\subsection{Synthesis of Statistical Findings}

Across both datasets and both statistical frameworks, three consistent findings provide empirically grounded guidance for pipeline configuration.

First, selecting a targeted number of initial and terminal points per trajectory end is significantly more effective than using all trajectory points. On the Bengaluru dataset, any value of $s$ in $\{1, 3, 5, 7, 9\}$ suffices; on UA-DETRAC, $s \in \{7, 9\}$ forms the best-performing group, with $s = 5$ not significantly different from either this group or $s = 3$. The value $s = 7$ consistently achieves the lowest or near-lowest error ranks across both datasets and both frameworks. This is consistent with the reasoning that interior trajectory points are spatially diffuse across the road surface and that their inclusion expands cluster extents, making adjacent entry and exit regions harder to separate.

Second, K-Means consistently outperforms GMM and DBSCAN, a result that is significant in every analysis conducted. The interpretation of this finding is discussed in Section~\ref{sec:conclusion}.

Third, the exclusion zone should impose a moderate buffer. Ratios between $1/4$ and $1/6$ achieve the lowest error across both datasets. Narrower buffers ($n \ge 7$) remove genuine boundary points; the absence of an exclusion zone, while not significantly worse than moderate ratios on either dataset individually, never achieves the lowest rank.

\begin{figure}[!htbp]
    \centering
    \includegraphics[width=\linewidth]{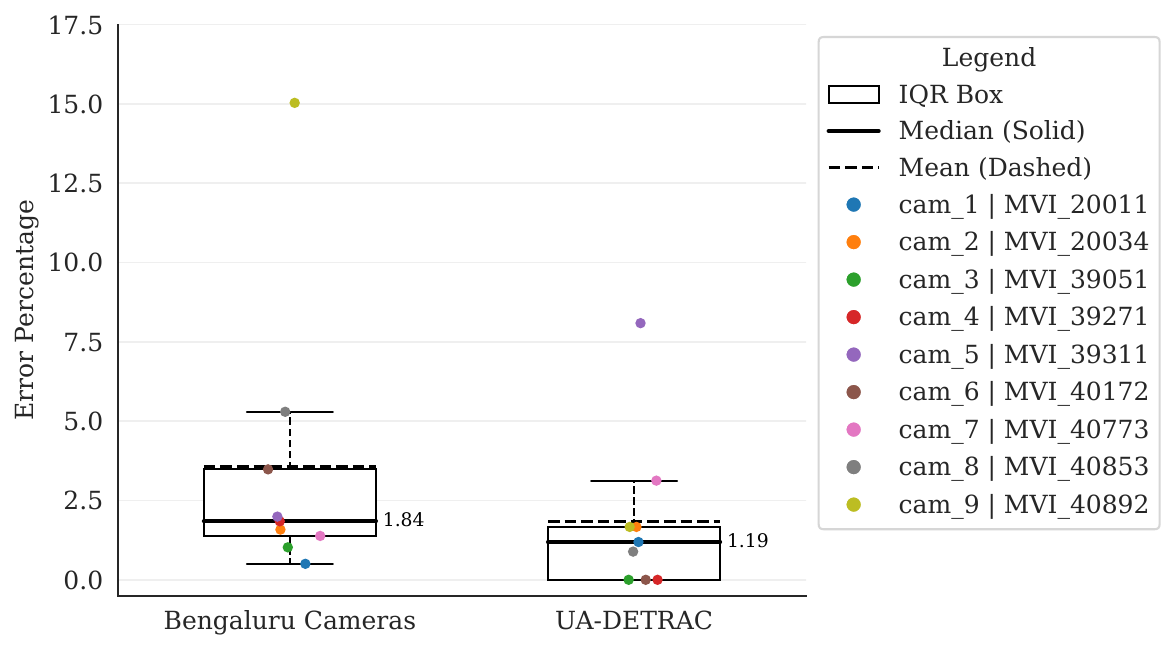}
    \caption{Classification error under the recommended configuration on the Bengaluru development cameras and UA-DETRAC sequences (train set).}
    \label{fig:train_recommended}
\end{figure}

The parameters that diverge in significance, ROI estimation, density equalization, and polygon representation, are those most sensitive to data quality and camera geometry. Their divergence is attributable to the distinct characteristics of the two datasets rather than to pipeline instability. The recommended configuration for subsequent evaluation uses $s = 7$ initial and terminal points per end, K-Means clustering, an exclusion ratio of $1/6$, ROI estimation enabled, no subsampling, and grid-based polygon representation. The exclusion ratio of $1/6$ is selected because the Bengaluru dataset represents the target deployment condition (dense heterogeneous traffic from uncalibrated surveillance cameras), whereas UA-DETRAC serves as a cross-domain sanity check; where the two datasets diverge, the Bengaluru-optimal setting is preferred. Of these six choices, the first three are supported by consistent statistical significance across both datasets; the latter three are supported by directional evidence and are adopted as reasonable defaults rather than as strongly validated optima. Fig.~\ref{fig:train_recommended} shows the performance of this configuration on both datasets, confirming that it generalizes across the two traffic regimes. This configuration is used in all subsequent experiments.

\subsection{Overall Performance Under Recommended Configuration}

\begin{figure}[!htbp]
    \centering
    \includegraphics[width=\linewidth]{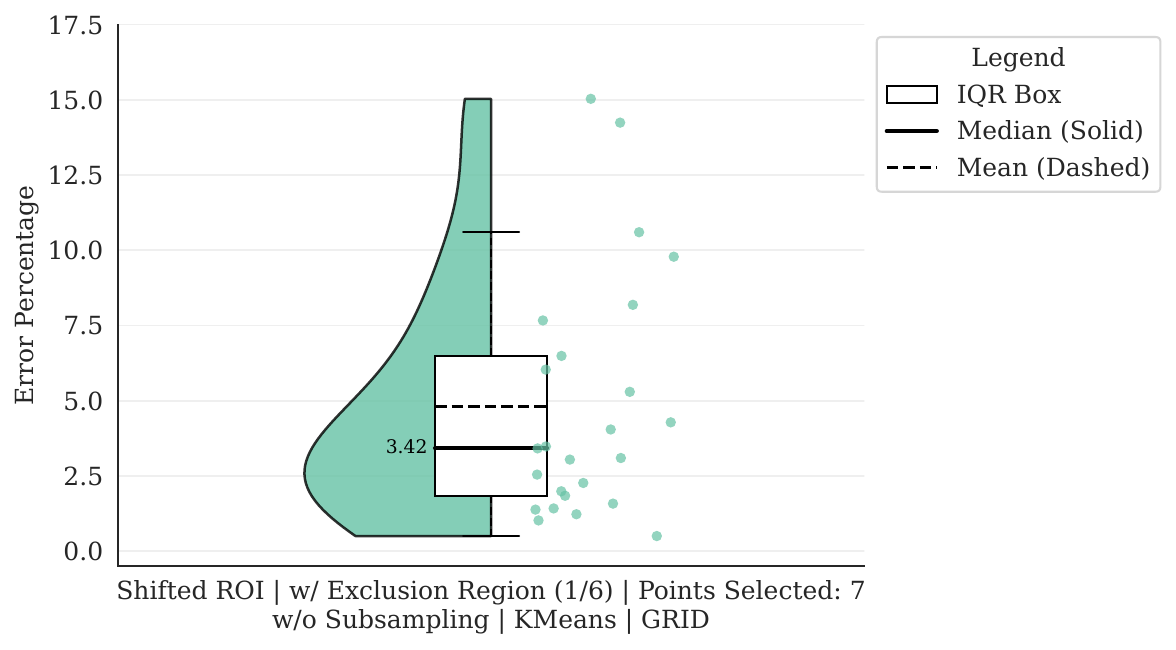}
    \caption{Distribution of classification error across all 25 camera locations under the recommended configuration. The 9 development locations and 16 held-out locations are shown together (test set).}
    \label{fig:test}
\end{figure}

The recommended configuration was applied to all 25 Bengaluru camera locations using the temporally independent video described in Section~\ref{subsec:dataset}, comprising the 9 development locations and 16 held-out locations. Fig.~\ref{fig:test} shows the distribution of classification error across all 25 locations. The pipeline achieves a median classification error of 3.42\% with a narrow interquartile range, indicating consistent performance across camera views that vary in mounting angle, field of view, intersection geometry, and traffic density.

\begin{figure}[!htbp]
    \centering
    \includegraphics[width=\linewidth]{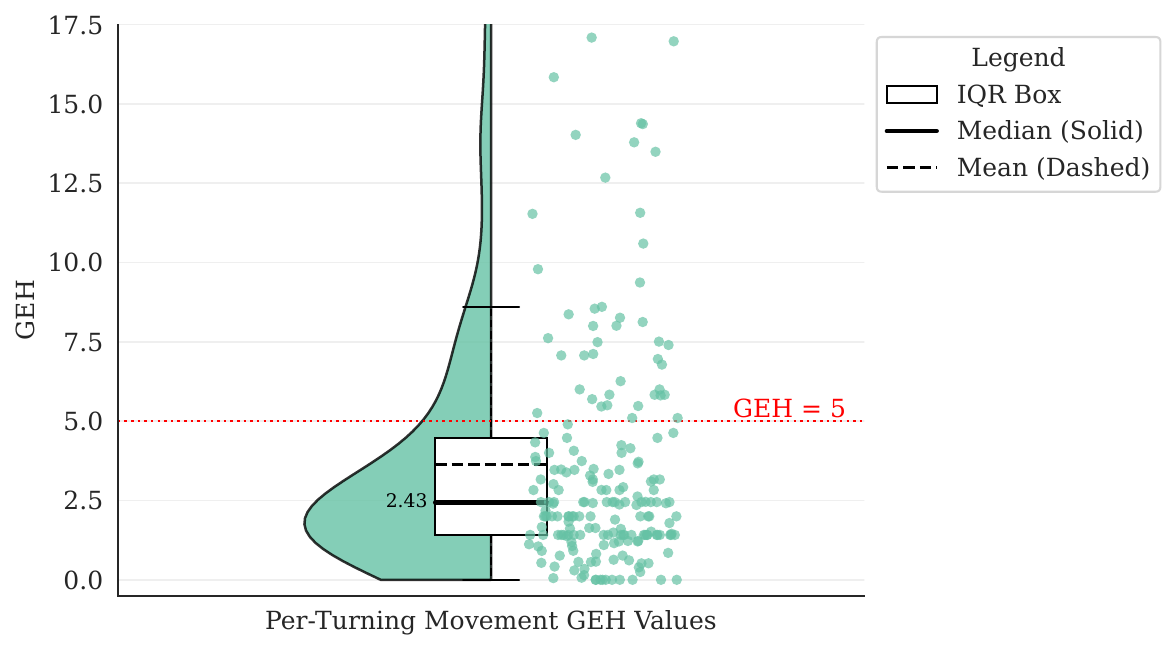}
    \caption{Distribution of per-turning-movement GEH values across all 25 camera locations under the recommended configuration (test set). The dashed red line indicates the GEH~$= 5$ threshold.}
    \label{fig:GEH}
\end{figure}

To situate these results within established practice in transportation engineering, the GEH statistic (Geoffrey E.\ Havers statistic)~\cite{Feldman2012} is additionally computed. The GEH statistic compares modeled and observed traffic flows using a formula that accounts for both absolute and relative differences, avoiding the distortion that percentage error exhibits at low flow volumes:
\begin{equation}
    \text{GEH} = \sqrt{\frac{2(M - C)^2}{M + C}},
\end{equation}
where $M$ is the predicted flow count and $C$ is the observed (ground-truth) flow count for a given turning movement. A GEH value below 5 is generally considered acceptable for individual turning movements at local roadway facilities~\cite{TxDOT2024Calibration}. The standard engineering criterion requires that at least 85\% of individual turning movements achieve GEH~$< 5$~\cite{TxDOT2024Calibration}. Fig.~\ref{fig:GEH} shows the distribution of per-turning-movement GEH values across all 25 camera locations: the median GEH is 2.43, and 76.77\% of individual turning movements achieve GEH~$< 5$. While this falls short of the 85\% criterion, the shortfall is concentrated in a small number of high-GEH outlier movements (range 0.0--19.2), consistent with the macro-averaging limitation discussed in Section~\ref{sec:conclusion} and likely attributable to upstream detection and tracking errors on rare or heavily occluded movements.

The trajectory assignment error and the GEH statistic evaluate complementary aspects of performance: the former measures whether individual trajectories are correctly classified, while the latter measures whether aggregate counts per movement match the ground truth. Since a pipeline could in principle achieve a low GEH value despite moderate assignment error if misclassifications are approximately symmetric, the fact that both metrics indicate strong central tendency strengthens the overall finding.

\subsection{Comparison with Trajectory Clustering Baselines}

To contextualize the proposed pipeline's performance, two trajectory clustering approaches were applied to the same 25 camera feeds and temporally independent video: the Hausdorff distance with DBSCAN from Jana et al.~\cite{Jana2023} and the LCSS (Longest Common Subsequence) similarity with K-Means from B\'{e}lisle et al.~\cite{Belisle2017}. These baselines were integrated into the pipeline after Step~2 (ROI estimation), with their trajectory clustering replacing Step~5. After clustering, the entry and exit points belonging to each clustered trajectory were identified and split into separate entry and exit clusters. Steps~3 (point selection) and~4 (exclusion zone) were then applied to these clusters, followed by Step~6 (polygon construction). The proposed pipeline's own preprocessing (Step~1) and ROI estimation (Step~2) were also applied, ensuring a controlled comparison that isolates the effect of the clustering approach.

\begin{figure}[!htbp]
    \centering
    \includegraphics[width=\linewidth]{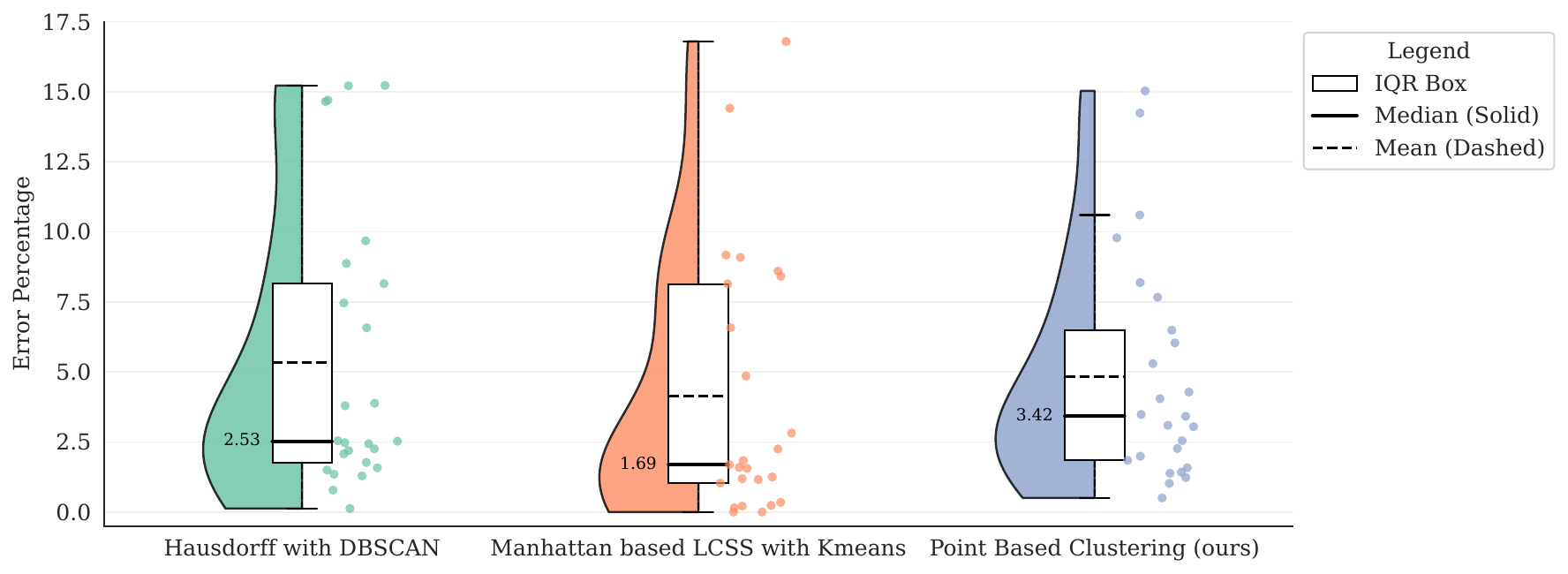}
    \caption{Classification error distribution across all 25 Bengaluru camera locations for the proposed pipeline (point clustering) and two trajectory clustering baselines. The proposed pipeline achieves the narrowest interquartile range despite higher median error (test set).}
    \label{fig:traj_clustering}
\end{figure}

Fig.~\ref{fig:traj_clustering} presents the error distributions for all three methods. Both trajectory clustering baselines achieve lower median classification error than the proposed pipeline. This outcome is expected: pairwise trajectory similarity measures (Hausdorff distance, LCSS) encode richer geometric information about vehicle paths than the initial and terminal point locations used by the proposed pipeline. However, three considerations qualify this comparison.

First, the proposed pipeline exhibits the lowest interquartile range among the three methods (Fig.~\ref{fig:traj_clustering}), indicating greater stability across the 25 camera views. In an operational deployment where the pipeline must perform reliably across thousands of cameras without per-camera tuning, predictable variance is a more relevant criterion than minimum median error on a curated set.

Second, the computational cost differs by complexity class. The proposed pipeline classifies each trajectory in $O(1)$ via point-in-polygon containment against persistent region polygons, yielding $O(N)$ total cost for $N$ trajectories. The trajectory clustering baselines require pairwise distance computation between all trajectory pairs, incurring $O(N^2)$ cost per clustering pass. As trajectory volumes increase with longer calibration periods or denser traffic, this quadratic scaling becomes the dominant computational bottleneck. In heavy traffic, a single 15-minute clip can contain thousands of trajectories; for the network-scale deployment envisioned in Section~\ref{sec:intro}, re-clustering every batch at $O(N^2)$ cost is impractical.

Third, the two approaches differ architecturally in a way that affects deployment. The proposed pipeline produces persistent spatial region polygons that, once estimated during a calibration period, classify any future trajectory by containment without re-computation. Trajectory clustering methods must be re-executed on every new batch of trajectories, as their output is a per-trajectory movement label rather than a reusable spatial structure. This distinction makes the proposed approach better suited to continuous monitoring, where regions are estimated once and applied indefinitely (subject to periodic recalibration as evaluated in Section~\ref{sec:extra_eval}).

In summary, the trajectory clustering baselines achieve higher accuracy on this evaluation set, while the proposed pipeline offers greater stability, lower computational cost, and a persistent region representation suited to network-scale deployment. The choice between approaches depends on whether the deployment priority is per-intersection accuracy or scalable, low-maintenance operation across large camera networks.

%%%%%%%%%%%%%%%%%%%%%%%%%%%%%%%%%%%%%%%%%%%%%%%%%%%%%%%%%%%%%%%%%%%%%%%%%%%%%%%%%%%%%%%%%%%%%%%%%%%%
%%%%%%%%%%%%%%%%%%%%%%%%%%%%%%%%%%%%%%%%%%%%%%%%%%%%%%%%%%%%%%%%%%%%%%%%%%%%%%%%%%%%%%%%%%%%%%%%%%%%
\section{Extended Evaluation}
\label{sec:extra_eval}

The parametric and nonparametric analyses identify which pipeline parameters matter most under controlled conditions. This section examines whether the recommended configuration remains effective under the variability encountered in operational deployment. Two factors are investigated, each addressing a practical question: how long should the calibration video be, and does traffic density during calibration matter? These experiments use the nine development camera locations, each with three days of video data providing 2:45 to 3:15 hours of footage per day. In all experiments, entry and exit regions are estimated from one day's video data and evaluated on non-overlapping clips from the remaining days, ensuring that the reported errors reflect the pipeline's ability to generalize beyond the trajectories used for region estimation.

\subsection{Effect of Calibration Duration}

\begin{figure}[!htbp]
    \centering
    \includegraphics[width=\linewidth]{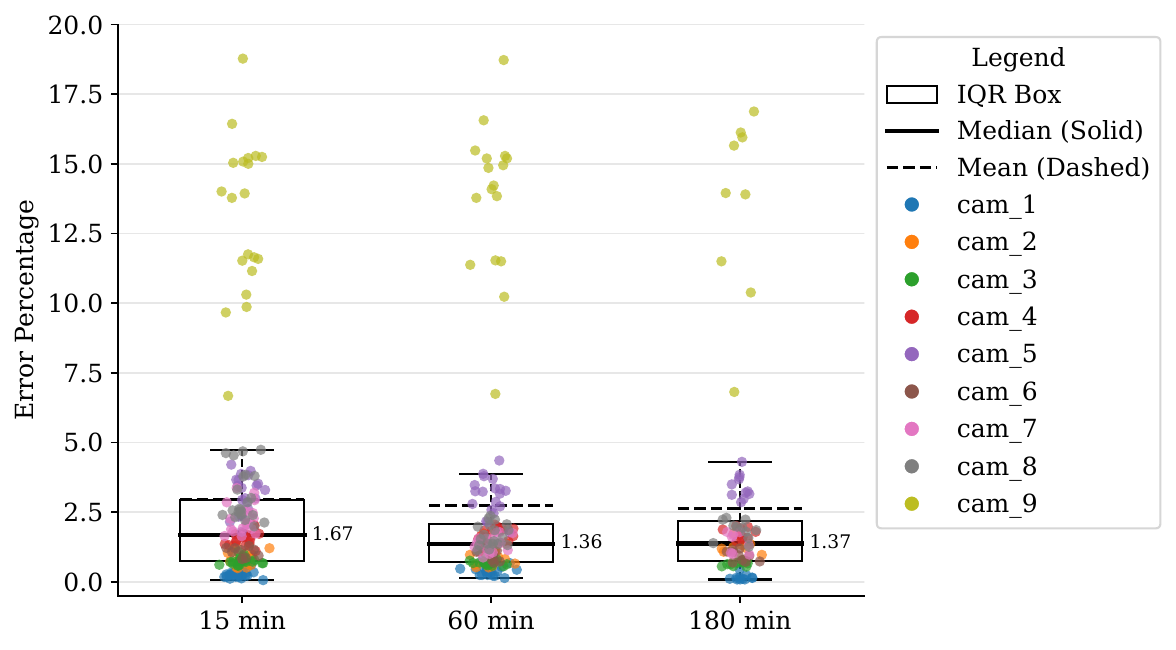}
    \caption{Distribution of per-movement classification error across 15-minute, 60-minute, and 180-minute calibration clips at all nine development camera locations. Each point represents the error for a single turning movement evaluated on a 15-minute test clip, color-coded by camera location.}
    \label{fig:treatment1}
\end{figure}

\emph{How does the duration of the calibration video affect region estimation quality?} Trajectories were extracted from 15-minute, 60-minute, and 180-minute clips at each of the nine development camera locations, and the pipeline was applied using the recommended configuration. As shown in Fig.~\ref{fig:treatment1}, longer calibration durations yield lower mean and median classification error, consistent with the expectation that a 15-minute clip may not capture the full variety of turning movements, particularly infrequent maneuvers such as U-turns. However, the improvement plateaus between the 60-minute and 180-minute conditions, showing little additional benefit beyond one hour. Where operationally feasible, using calibration clips of at least 60 minutes improves the generalizability of the estimated regions without requiring substantially longer video.

\subsection{Effect of Traffic Density}

\begin{figure}[!htbp]
    \centering
    \includegraphics[width=\linewidth]{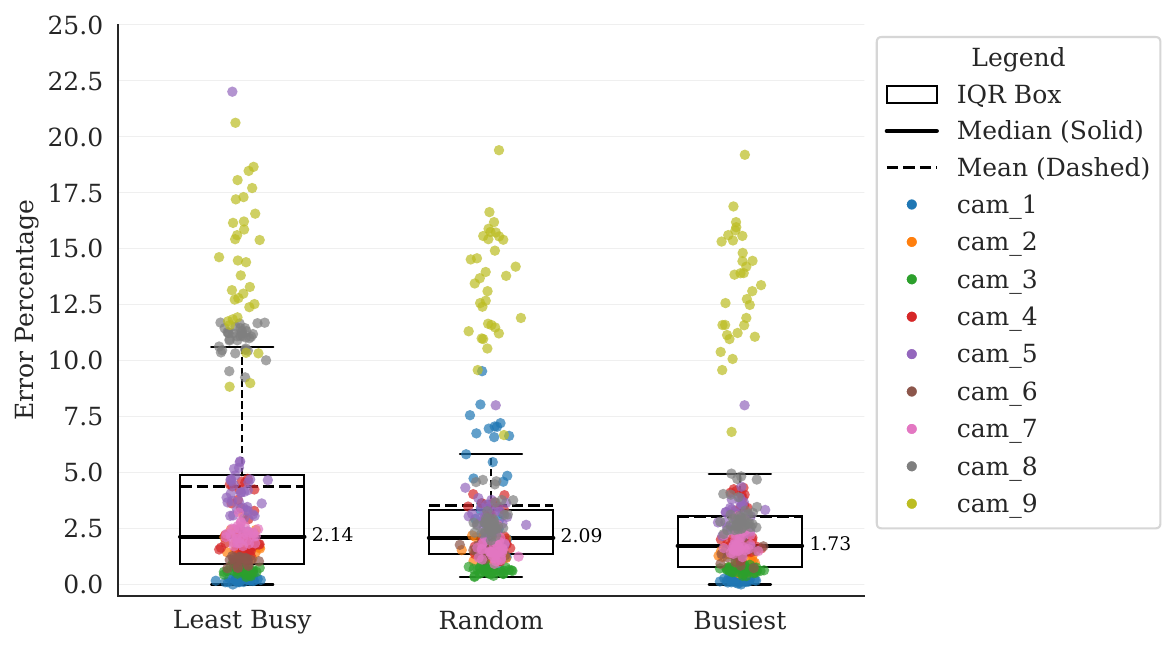}
    \caption{Distribution of classification error across the least busy, randomly selected, and busiest 15-minute calibration clips at all nine development camera locations. Each point represents the error for a single turning movement evaluated on a 15-minute test clip, color-coded by camera location.}
    \label{fig:treatment3}
\end{figure}

\emph{Does selecting a high-density calibration clip improve region estimation compared to a randomly chosen or low-density clip?} Regions were generated from the busiest 15-minute clip at each location (highest trajectory count), a randomly selected 15-minute clip (selected once per camera), and the least busy 15-minute clip (lowest trajectory count with at least 50 trajectories). As shown in Fig.~\ref{fig:treatment3}, the high-density condition yields substantially lower error. During congestion, vehicles traverse the full range of entry and exit regions, including minor approach roads and less frequent turns, providing the clustering step with more complete spatial coverage. The least busy condition produces the highest error, as low-volume clips may omit entire turning movements. When only a short calibration clip is available, selecting one from a peak-traffic period is likely to yield the most representative regions.

%%%%%%%%%%%%%%%%%%%%%%%%%%%%%%%%%%%%%%%%%%%%%%%%%%%%%%%%%%%%%%%%%%%%%%%%%%%%%%%%%%%%%%%%%%%%%%%%%%%%
%%%%%%%%%%%%%%%%%%%%%%%%%%%%%%%%%%%%%%%%%%%%%%%%%%%%%%%%%%%%%%%%%%%%%%%%%%%%%%%%%%%%%%%%%%%%%%%%%%%%
\section{Discussion and Conclusion}
\label{sec:conclusion}

This paper presents an unsupervised pipeline that identifies entry and exit regions directly from raw vehicle trajectory data, enabling turning movement classification without manual region annotation across diverse camera views. Systematic evaluation of six pipeline parameters using both parametric and nonparametric statistical frameworks yields three consistent findings, alongside several deployment-relevant observations.

\subsection{Key Findings}

As established in Section~\ref{sec:results}, three pipeline parameters consistently and significantly affect classification accuracy across both datasets and both statistical frameworks: point selection, clustering algorithm, and exclusion zone ratio.

The superiority of selecting $s = 7$ or $s = 9$ points per end over using all trajectory points is consistent with the spatial structure of the data: interior points encode path information rather than boundary location, and their inclusion expands each cluster's spatial extent, making adjacent entry and exit regions harder to separate. This effect is particularly pronounced in cameras mounted at low height or steep pitch, where the projected distance between adjacent regions is already small.

The consistent advantage of K-Means over GMM and DBSCAN can be attributed to implicit regularization. K-Means functions as a highly constrained variant of GMM, assuming isotropic, uniformly sized clusters. Although entry and exit regions are rarely perfectly spherical, this geometric constraint serves as a structural safeguard: rather than adapting to absorb noisy, scattered trajectories, K-Means enforces tight, compact boundaries that prevent adjacent entry and exit zones from merging. DBSCAN, by contrast, is sensitive to the density variations that arise from both perspective projection and differences in vehicle speed near intersections, complicating the selection of a single global density threshold.

The exclusion zone's effectiveness depends on the ROI estimated in Step~2: when the camera is oriented toward the horizon, entry and exit points may appear near the center of the image frame, and estimating the ROI from observed vehicle activity ensures that the exclusion zone corresponds to the actual intersection interior. The recommended configuration therefore enables ROI estimation, and the exclusion zone is defined relative to this estimated region. Perspective density equalization does not improve performance on either dataset, suggesting that K-Means is sufficiently robust to moderate density gradients and that the information lost through subsampling outweighs any equalization benefit.

Under the recommended configuration, the pipeline achieves a median classification error of 3.42\% across all 25 camera locations, including 16 held-out feeds. The median per-turning-movement GEH is 2.43~\cite{Feldman2012}, though 76.77\% of individual turning movements achieve GEH~$< 5$, falling short of the 85\% engineering criterion~\cite{TxDOT2024Calibration}; the shortfall is attributable to a small number of high-GEH outlier movements likely driven by upstream detection and tracking errors.

\subsection{Limitations}

Several limitations should be acknowledged. The pipeline's accuracy is contingent on the quality of upstream detection and tracking; severe and persistent occlusion can produce trajectory data of insufficient quality for reliable region estimation.

Temporal robustness over extended deployment periods (weeks or months) has not been assessed. Over such timescales, gradual physical displacement of the camera due to wind, vibration, or mounting degradation (commonly termed camera drift) can alter the field of view, and environmental changes such as vegetation growth or construction may modify the road layout. Either condition could invalidate the estimated regions, requiring recalibration.

The fixed preprocessing thresholds, including the spatial displacement threshold $\delta_{min}$, were set empirically for $1920 \times 1080$ resolution and have not been validated at other resolutions.

The ground-truth annotations were produced through expert consensus. Although annotators discussed and resolved disagreements, formal inter-annotator agreement metrics (such as Cohen's kappa) were not computed prior to consensus resolution.

The evaluation metric uses a macro-average across turning movements, weighting each movement equally regardless of volume. For rare movements (e.g., U-turns with very few trajectories), $TN_{(o,d)}$ dominates the denominator, and $E_{(o,d)}$ can appear low even when most trajectories of that movement are misclassified. The reported median errors may therefore understate classification difficulty for low-volume movements.

The statistical analysis evaluates each pipeline parameter independently via one-way tests. Interactions between parameters, such as the documented coupling between the exclusion ratio and grid resolution, are not modeled. A factorial or mixed-effects analysis could reveal interaction effects that the current framework does not capture.

\subsection{Future Work}

Several directions follow from these findings. Extending evaluation to multi-week deployment periods would address the temporal robustness gap and enable investigation of adaptive region-updating strategies that respond to detected changes in the scene.

Incorporating vehicle class information could enable class-specific turning movement classification, which is valuable in heterogeneous traffic environments where movement patterns differ across vehicle types (for example, heavy vehicles may be restricted to certain turns).

The pipeline currently operates in the image plane and is therefore subject to perspective distortion, which the exclusion zone and optional density equalization partially address. An alternative approach would be to project trajectories into a bird's-eye view (BEV) or orthographic representation prior to clustering. BEV projection has been shown to improve spatial reasoning in traffic scenes~\cite{Pakdamansavoji2024}; however, it requires either known camera intrinsic and extrinsic parameters or a learned homography, both of which conflict with the camera-agnostic requirement that motivates the present work. Investigating calibration-free BEV approximation methods that preserve the unsupervised, camera-agnostic design is a promising direction.

Developing resolution-independent methods to compensate for perspective distortion, potentially through learned depth-aware weighting, could further improve clustering quality without requiring camera-specific parameters.

Finally, integrating the proposed pipeline with downstream applications such as adaptive signal control and origin-destination estimation would demonstrate its utility within the broader intelligent transportation ecosystem.

%%%%%%%%%%%%%%%%%%%%%%%%%%%%%%%%%%%%%%%%%%%%%%%%%%%%%%%%%%%%%%%%%%%%%%%%%%%%%%%%%%%%%%%%%%%%%%%%%%%%
%%%%%%%%%%%%%%%%%%%%%%%%%%%%%%%%%%%%%%%%%%%%%%%%%%%%%%%%%%%%%%%%%%%%%%%%%%%%%%%%%%%%%%%%%%%%%%%%%%%%
\section*{Acknowledgments}
The authors gratefully acknowledge the support and resources provided by Centre for infrastructure, Sustainable Transportation \& Urban Planning (CiSTUP) @ IISc; AI \& Robotics Technology Park (ARTPARK), I-Hub @ IISc;  Arcot Ramachandran Young Investigator Award from IISc; India Urban Data Exchange (IUDX); and the Bengaluru Traffic Police (BTP), which made this research possible.

\bibliographystyle{IEEEtran}
\bibliography{references.bib}

\clearpage
\onecolumn

% Reset counters for supplementary material
\setcounter{page}{1}
\setcounter{section}{0}
\setcounter{figure}{0}
\setcounter{table}{0}
\setcounter{equation}{0}
\setcounter{algorithm}{0}
\renewcommand{\thesection}{S-\Roman{section}}
\renewcommand{\thefigure}{S\arabic{figure}}
\renewcommand{\thetable}{S\arabic{table}}
\renewcommand{\theequation}{S\arabic{equation}}

\begin{center}
    \Large\bfseries Supplementary Material\\[6pt]
    \normalsize Unsupervised Detection of Entry and Exit Regions from Vehicle Trajectories\\
    for Camera-Agnostic Turning Movement Counts
\end{center}
\vspace{1em}

%%%%%%%%%%%%%%%%%%%%%%%%%%%%%%%%%%%%%%%%%%%%%%%%%%%%%%%%%%%%%%%%%%%%%%%%%%%%%%%%%%%%%%%%%%%%%%%%%%%%
\section{Preprocessing Threshold Selection}
\label{supp:thresholds}

The trajectory preprocessing step (Section~III, Step~1 in the main text) applies three sequential filters, each governed by a threshold. The values used across all experiments are as follows.

\begin{figure}[htbp]
    \centering
    \begin{minipage}[t]{0.32\textwidth}
        \centering
        \includegraphics[width=\linewidth]{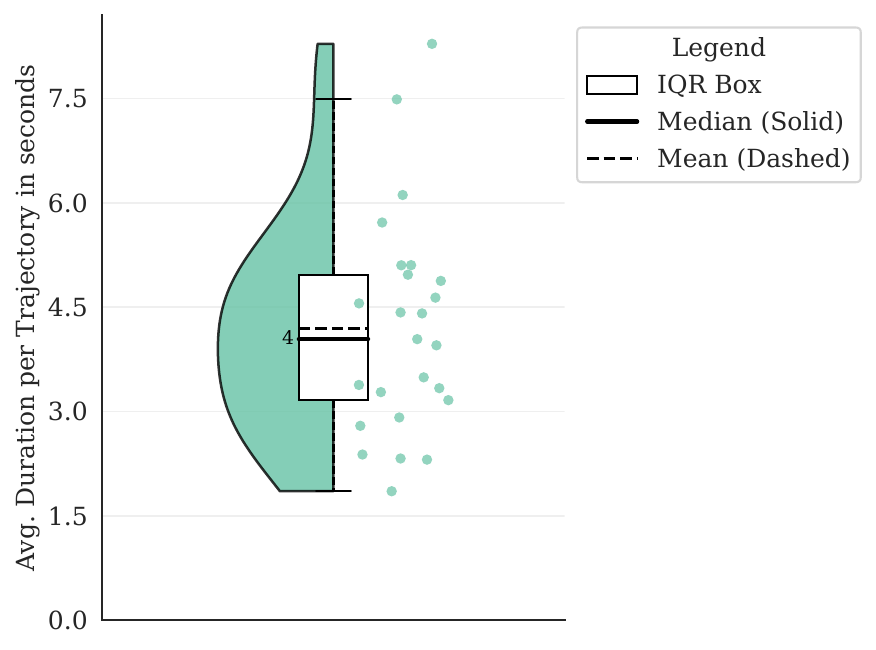}
        \caption{Distribution of average trajectory duration across cameras.}
        \label{fig:supp_duration}
    \end{minipage}
    \hfill
    \begin{minipage}[t]{0.32\textwidth}
        \centering
        \includegraphics[width=\linewidth]{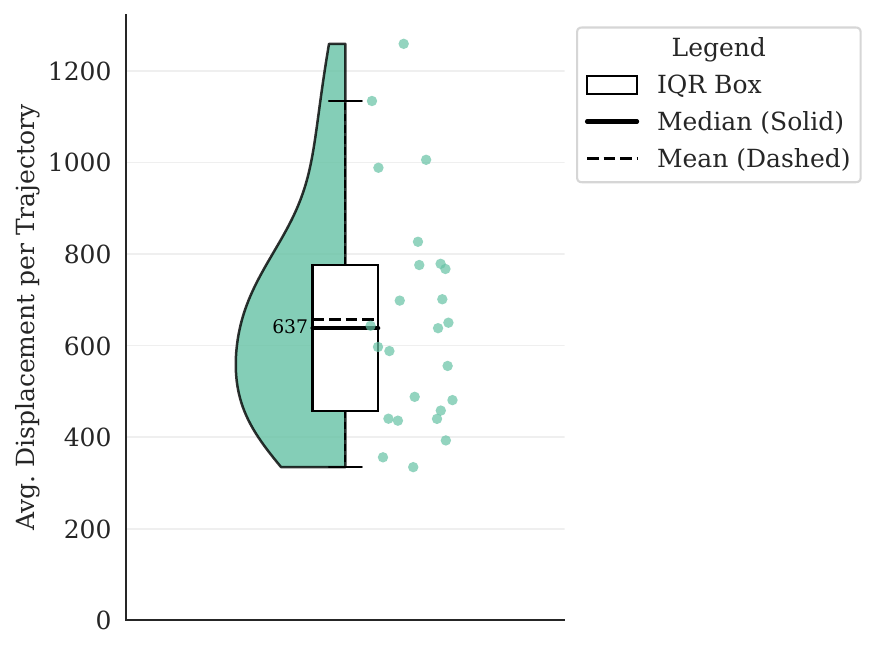}
        \caption{Distribution of average trajectory displacement across cameras.}
        \label{fig:supp_displacement}
    \end{minipage}
    \hfill
    \begin{minipage}[t]{0.32\textwidth}
        \centering
        \includegraphics[width=\linewidth]{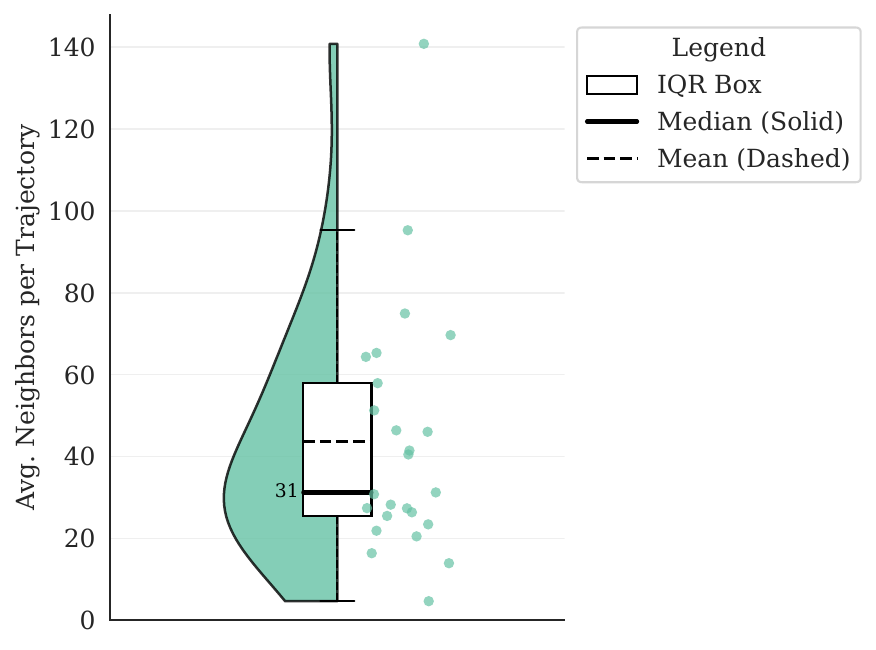}
        \caption{Distribution of average neighbor count per trajectory across cameras (radius $r = 50$~pixels).}
        \label{fig:supp_neighbors}
    \end{minipage}
\end{figure}

\textbf{Minimum trajectory duration} ($\tau_{min} = 1.5$~seconds). At 30~FPS, for example, this corresponds to 45~frames. Fig.~\ref{fig:supp_duration} shows the distribution of average trajectory duration across all camera locations; the vast majority of trajectories exceed this threshold, confirming that it removes only short fragments likely caused by identity switches.

\textbf{Minimum displacement} ($\delta_{min} = 72$~pixels). This value corresponds to approximately 3.3\% of the image diagonal at $1920 \times 1080$ resolution. It was set empirically to exclude stationary or near-stationary vehicles (e.g., parked cars). Fig.~\ref{fig:supp_displacement} shows the distribution of average displacement across cameras; the threshold falls well below the median, affecting only a small number of trajectories. Validating this threshold across other resolutions remains a direction for future work.

\textbf{Minimum neighbor count} ($n_{min} = 3$, radius $r = 50$~pixels). A trajectory is retained only if both its initial and terminal points have at least $n_{min} = 3$ distinct neighbors within radius $r$ in the global pool of initial and terminal points. Fig.~\ref{fig:supp_neighbors} shows the distribution of average neighbor counts; the threshold of 3 is substantially below the median of 31, ensuring that only genuinely isolated points are removed.

\begin{figure}[htbp]
    \centering
    \includegraphics[width=\textwidth]{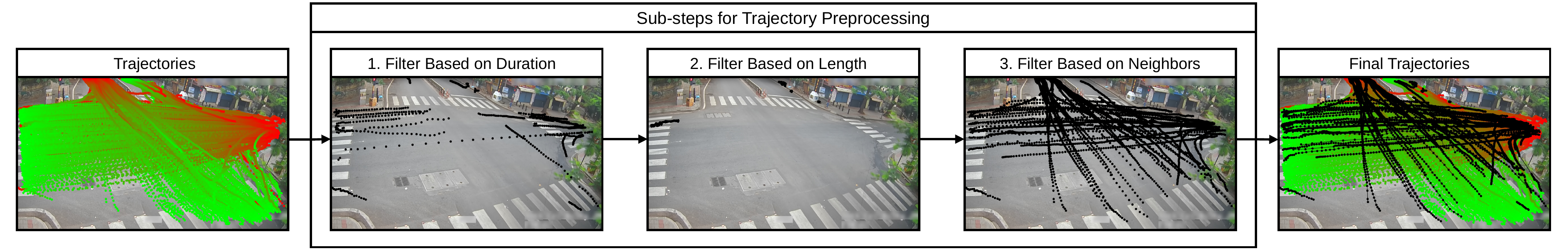}
    \caption{Three-stage trajectory preprocessing pipeline applied to a representative camera feed (see Step~1 in Fig.~1 of the main text). Each stage progressively removes spurious trajectories.}
    \label{fig:supp_preprocess_pipeline}
\end{figure}

Fig.~\ref{fig:supp_preprocess_pipeline} illustrates the cumulative effect of the three filters on a representative camera feed: of 102 raw trajectories, 32 were removed by the temporal filter, 9 by the displacement filter, and 61 by the neighborhood filter, yielding a filtered set that retains trajectories corresponding to genuine vehicle movements.

%%%%%%%%%%%%%%%%%%%%%%%%%%%%%%%%%%%%%%%%%%%%%%%%%%%%%%%%%%%%%%%%%%%%%%%%%%%%%%%%%%%%%%%%%%%%%%%%%%%%
\section{Pipeline Pseudocode}
\label{supp:pseudocode}

Algorithm~\ref{alg:supp_pipeline} provides the complete pseudocode for the proposed pipeline. The procedure takes the raw trajectory set $\mathcal{T}_{raw}$ and an input image $I$ as inputs and returns a set of region polygons $\mathcal{H}$ delineating the identified entry and exit regions. Note that clustering is performed separately on initial points and terminal points (line~16); for brevity, the pseudocode shows a single pass, which is executed once for each point type.

\begin{algorithm}[htbp]
\caption{Entry and Exit Region Polygon Generation}
\label{alg:supp_pipeline}
\begin{algorithmic}[1]

\Require Raw trajectory set $\mathcal{T}_{raw}$, input image $I$, exclusion ratio $1/n$, points per end $s$
\Ensure Set of region polygons $\mathcal{H}$

\Procedure{GenerateRegions}{$\mathcal{T}_{raw}, I, n, s$}

    \State $\mathcal{T}_{filtered} \gets \Call{PreprocessTrajectories}{\mathcal{T}_{raw}, \tau_{min}, \delta_{min}, n_{min}, r}$
    \Comment{Step 1: temporal, displacement, and neighborhood filtering}

    \State $\mathcal{R} \gets \Call{ComputeROI}{\mathcal{T}_{filtered}}$
    \Comment{Step 2: bounding rectangle of all trajectory coordinates}

    \State $\mathcal{P}_{selected} \gets \emptyset$
    \For{\textbf{each} trajectory $T \in \mathcal{T}_{filtered}$}
        \State $\mathcal{P}_{selected} \gets \mathcal{P}_{selected} \cup \Psi(T, s)$
        \Comment{Step 3: select $s$ initial and $s$ terminal points}
    \EndFor

    \State $\mathcal{P}_{equalized} \gets \Call{PerspectiveEqualize}{\mathcal{P}_{selected}}$
    \Comment{Step 3 (optional): density equalization}

    \State $\mathcal{R}_{excl} \gets \Call{ComputeExclusionZone}{\mathcal{R}, n}$
    \Comment{Step 4: concentric rectangle with margins $W_\mathcal{R}/n$, $H_\mathcal{R}/n$}
    \State $\mathcal{P}_{boundary} \gets \{ p \in \mathcal{P}_{equalized} \mid p \notin \mathcal{R}_{excl} \}$

    \State \Comment{Step 5: cluster initial and terminal points separately}
    \State $\mathcal{C}_{entry} \gets \Call{Cluster}{\{p \in \mathcal{P}_{boundary} \mid p \text{ is an initial point}\}}$
    \State $\mathcal{C}_{exit} \gets \Call{Cluster}{\{p \in \mathcal{P}_{boundary} \mid p \text{ is a terminal point}\}}$

    \State \Comment{Step 6: construct region polygons}
    \State $\mathcal{H} \gets \emptyset$
    \For{\textbf{each} cluster $C_j \in \mathcal{C}_{entry} \cup \mathcal{C}_{exit}$}
        \State $H_j \gets \Call{ConvexHull}{C_j}$
        \State $H_j \gets H_j \setminus \mathcal{R}_{excl}$
        \Comment{Clip hull to buffer zone}
        \State $\mathcal{H} \gets \mathcal{H} \cup \{H_j\}$
    \EndFor

    \State \Comment{Optional: grid-based expansion}
    \State $\mathcal{G} \gets \Call{DivideIntoGrid}{\mathcal{R}, n}$
    \Comment{$n \times n$ grid over ROI}
    \For{\textbf{each} cell $g_i \in \mathcal{G}$ with $g_i \subseteq \mathcal{R} \setminus \mathcal{R}_{excl}$}
        \State $j^* \gets \operatorname*{argmax}_{j} |\mathcal{P}_{boundary} \cap g_i \cap H_j|$
        \If{$|\mathcal{P}_{boundary} \cap g_i| > 0$}
            \State Assign $g_i$ to region $R_{j^*}$
        \EndIf
    \EndFor
    \State $\mathcal{H} \gets \Call{MergeGridCells}{\mathcal{H}, \mathcal{G}}$
    \Comment{Merge cells into region polygons}

    \State \Return $\mathcal{H}$

\EndProcedure

\end{algorithmic}
\end{algorithm}

%%%%%%%%%%%%%%%%%%%%%%%%%%%%%%%%%%%%%%%%%%%%%%%%%%%%%%%%%%%%%%%%%%%%%%%%%%%%%%%%%%%%%%%%%%%%%%%%%%%%
\section{Time Complexity Analysis}
\label{supp:time_complexity}

Table~\ref{tab:supp_complexity} summarizes the computational complexity of each pipeline step. Let $N$ denote the number of trajectories, $L$ the average trajectory length, and $s$ the number of selected points per end.

\begin{table}[htbp]
\centering
\caption{Per-step computational complexity of the proposed pipeline. $N$: number of trajectories; $L$: average trajectory length; $s$: points per end; $K$: number of clusters; $n$: exclusion ratio denominator.}
\label{tab:supp_complexity}
\begin{tabular}{@{}llc@{}}
\toprule
\textbf{Step} & \textbf{Operation} & \textbf{Complexity} \\
\midrule
1. Preprocessing & Temporal and displacement filters & $O(N)$ \\
                  & Neighborhood filter & $O(N \log N)$ \\
2. ROI Estimation & Coordinate extrema & $O(NL)$ \\
3. Point Selection & Extract $s$ points per end & $O(Ns)$ \\
4. Exclusion Zone & Point filtering & $O(Ns)$ \\
5. Clustering (K-Means) & $K$ iterations over boundary points & $O(KNs)$ \\
6. Polygon Construction & Convex hull per cluster & $O(Ns \log Ns)$ \\
                        & Grid cell assignment & $O(n^2)$ \\
\midrule
\textbf{Total (region estimation)} & & $O(NL)$ \\
\midrule
\textbf{Classification (per trajectory)} & Point-in-polygon containment & $O(1)$ \\
\textbf{Classification ($N$ trajectories)} & & $O(N)$ \\
\bottomrule
\end{tabular}
\end{table}

The dominant term in region estimation is the ROI computation, which scans all $NL$ trajectory coordinates. All subsequent steps operate on the $2Ns$ selected points, where $s$ is a small constant ($s = 7$ in the recommended configuration). Clustering with K-Means introduces a factor of $K$ (bounded between 2 and 9), which does not change the asymptotic complexity. The neighborhood filter in Step~1 uses a KD-tree for spatial queries, yielding $O(N \log N)$ rather than $O(N^2)$.

The critical distinction from trajectory clustering baselines lies in the classification phase. Once region polygons are estimated, the proposed pipeline classifies each new trajectory in $O(1)$ via point-in-polygon containment, yielding $O(N)$ total cost for a batch of $N$ trajectories. By contrast, trajectory clustering baselines (Jana et al.~\cite{Jana2023}, B\'{e}lisle et al.~\cite{Belisle2017}) require pairwise similarity computation between all trajectory pairs, incurring $O(N^2 L^2)$ cost per batch (the $L^2$ factor arises from Hausdorff distance or LCSS computation between trajectory pairs of length $L$). Region estimation is a one-time cost amortized over all subsequent classification; the baselines must re-execute the full $O(N^2 L^2)$ computation for every new batch.

%%%%%%%%%%%%%%%%%%%%%%%%%%%%%%%%%%%%%%%%%%%%%%%%%%%%%%%%%%%%%%%%%%%%%%%%%%%%%%%%%%%%%%%%%%%%%%%%%%%%
\section{Point Selection Function}
\label{supp:psi}

The selection function $\Psi(T, s)$ extracts the first and last $s$ points from each trajectory $T = (p_1, \dots, p_L)$, where $s$ is the number of points selected per end and $L$ is the trajectory length:
\begin{equation}
    \Psi(T, s) =
    \begin{cases}
        (p_1,\; p_L), & s = 1, \\[6pt]
        \begin{aligned}
        &(p_1, \dots, p_s) \;\cup \\
        &(p_{L-s+1}, \dots, p_L),
        \end{aligned}
        & 1 < s < \lfloor L/2 \rfloor, \\[6pt]
        T, & s \ge \lfloor L/2 \rfloor.
    \end{cases}
\end{equation}
When $s = 1$, the function returns the single initial and single terminal point. When $1 < s < \lfloor L/2 \rfloor$, two non-overlapping subsequences of length $s$ are extracted from the beginning and end of the trajectory. When $s \ge \lfloor L/2 \rfloor$, the two subsequences would overlap or exhaust the trajectory, and all points are retained. The values of $s$ evaluated in this study are $s \in \{1, 3, 5, 7, 9\}$ and $s \ge \lfloor L/2 \rfloor$ (all points).

%%%%%%%%%%%%%%%%%%%%%%%%%%%%%%%%%%%%%%%%%%%%%%%%%%%%%%%%%%%%%%%%%%%%%%%%%%%%%%%%%%%%%%%%%%%%%%%%%%%%
\section{Tukey HSD Pairwise Comparisons: Bengaluru Dataset}
\label{supp:tukey_blr}

Tables~\ref{tab:supp_tukey_roi_blr}--\ref{tab:supp_tukey_poly_blr} present the complete pairwise Tukey HSD post-hoc comparisons for each pipeline parameter on the Bengaluru dataset, expanding the summary in Table~II of the main text. Tables are ordered by pipeline step number for consistency with the main text. A family-wise error rate of $\alpha = 0.05$ was maintained. On this dataset, the only parameters yielding significant pairwise differences are point selection (any value of $s$ vs.\ using all points), exclusion zone ratio (all ratios vs.\ $1/8$, and $1/5$ vs.\ $1/7$), and clustering algorithm (all three pairs are significant, with K-Means achieving the lowest mean error).

\begin{table}[htbp]
\centering
\caption{Tukey HSD: ROI estimation (Bengaluru). Not significant.}
\label{tab:supp_tukey_roi_blr}
\begin{tabular}{@{}llrrrc@{}}
\toprule
\textbf{Group 1} & \textbf{Group 2} & \textbf{Mean Diff.} & \textbf{$p$-adj} & \textbf{95\% CI} & \textbf{Sig.} \\
\midrule
Without ROI & With ROI & $-0.035$ & 0.761 & $[-0.254, 0.184]$ & No \\
\bottomrule
\end{tabular}
\end{table}

\begin{table}[htbp]
\centering
\caption{Tukey HSD: Point selection, $s$ (Bengaluru). Significant differences ($p<0.05$) in bold.}
\label{tab:supp_tukey_points_blr}
\begin{tabular}{@{}llrrrc@{}}
\toprule
\textbf{Group 1} & \textbf{Group 2} & \textbf{Mean Diff.} & \textbf{$p$-adj} & \textbf{95\% CI} & \textbf{Sig.} \\
\midrule
$s=1$ & $s=3$ & $-0.115$ & 0.984 & $[-0.600, 0.369]$ & No \\
$s=1$ & $s=5$ & $0.076$ & 0.998 & $[-0.408, 0.561]$ & No \\
$s=1$ & $s=7$ & $-0.024$ & 1.000 & $[-0.509, 0.460]$ & No \\
$s=1$ & $s=9$ & $0.024$ & 1.000 & $[-0.461, 0.509]$ & No \\
$s=1$ & All & $2.580$ & \textbf{$<$0.001} & $[2.095, 3.065]$ & \textbf{Yes} \\
\midrule
$s=3$ & $s=5$ & $0.192$ & 0.870 & $[-0.293, 0.677]$ & No \\
$s=3$ & $s=7$ & $0.091$ & 0.995 & $[-0.394, 0.576]$ & No \\
$s=3$ & $s=9$ & $0.139$ & 0.964 & $[-0.345, 0.624]$ & No \\
$s=3$ & All & $2.695$ & \textbf{$<$0.001} & $[2.210, 3.180]$ & \textbf{Yes} \\
\midrule
$s=5$ & $s=7$ & $-0.101$ & 0.992 & $[-0.586, 0.384]$ & No \\
$s=5$ & $s=9$ & $-0.052$ & 1.000 & $[-0.537, 0.432]$ & No \\
$s=5$ & All & $2.503$ & \textbf{$<$0.001} & $[2.019, 2.988]$ & \textbf{Yes} \\
\midrule
$s=7$ & $s=9$ & $0.048$ & 1.000 & $[-0.436, 0.533]$ & No \\
$s=7$ & All & $2.604$ & \textbf{$<$0.001} & $[2.119, 3.089]$ & \textbf{Yes} \\
\midrule
$s=9$ & All & $2.556$ & \textbf{$<$0.001} & $[2.071, 3.041]$ & \textbf{Yes} \\
\bottomrule
\end{tabular}
\end{table}

\begin{table}[htbp]
\centering
\caption{Tukey HSD: Exclusion zone ratio (Bengaluru). Significant differences ($p<0.05$) in bold.}
\label{tab:supp_tukey_excl_blr}
\begin{tabular}{@{}llrrrc@{}}
\toprule
\textbf{Group 1} & \textbf{Group 2} & \textbf{Mean Diff.} & \textbf{$p$-adj} & \textbf{95\% CI} & \textbf{Sig.} \\
\midrule
None & 1/3 & $-0.178$ & 0.996 & $[-1.000, 0.644]$ & No \\
None & 1/4 & $-0.470$ & 0.624 & $[-1.292, 0.351]$ & No \\
None & 1/5 & $-0.558$ & 0.414 & $[-1.379, 0.264]$ & No \\
None & 1/6 & $-0.418$ & 0.746 & $[-1.239, 0.404]$ & No \\
None & 1/7 & $0.046$ & 1.000 & $[-0.776, 0.867]$ & No \\
None & 1/8 & $1.201$ & \textbf{$<$0.001} & $[0.380, 2.023]$ & \textbf{Yes} \\
\midrule
1/3 & 1/4 & $-0.292$ & 0.644 & $[-0.812, 0.227]$ & No \\
1/3 & 1/5 & $-0.380$ & 0.321 & $[-0.899, 0.140]$ & No \\
1/3 & 1/6 & $-0.240$ & 0.823 & $[-0.759, 0.280]$ & No \\
1/3 & 1/7 & $0.224$ & 0.866 & $[-0.296, 0.743]$ & No \\
1/3 & 1/8 & $1.379$ & \textbf{$<$0.001} & $[0.860, 1.899]$ & \textbf{Yes} \\
\midrule
1/4 & 1/5 & $-0.087$ & 0.999 & $[-0.607, 0.432]$ & No \\
1/4 & 1/6 & $0.053$ & 1.000 & $[-0.467, 0.572]$ & No \\
1/4 & 1/7 & $0.516$ & 0.053 & $[-0.003, 1.036]$ & No \\
1/4 & 1/8 & $1.672$ & \textbf{$<$0.001} & $[1.152, 2.191]$ & \textbf{Yes} \\
\midrule
1/5 & 1/6 & $0.140$ & 0.986 & $[-0.380, 0.660]$ & No \\
1/5 & 1/7 & $0.603$ & \textbf{0.011} & $[0.084, 1.123]$ & \textbf{Yes} \\
1/5 & 1/8 & $1.759$ & \textbf{$<$0.001} & $[1.239, 2.279]$ & \textbf{Yes} \\
\midrule
1/6 & 1/7 & $0.464$ & 0.117 & $[-0.056, 0.983]$ & No \\
1/6 & 1/8 & $1.619$ & \textbf{$<$0.001} & $[1.099, 2.139]$ & \textbf{Yes} \\
\midrule
1/7 & 1/8 & $1.155$ & \textbf{$<$0.001} & $[0.636, 1.675]$ & \textbf{Yes} \\
\bottomrule
\end{tabular}
\end{table}

\begin{table}[htbp]
\centering
\caption{Tukey HSD: Clustering algorithm (Bengaluru). All pairs are significant.}
\label{tab:supp_tukey_clust_blr}
\begin{tabular}{@{}llrrrc@{}}
\toprule
\textbf{Group 1} & \textbf{Group 2} & \textbf{Mean Diff.} & \textbf{$p$-adj} & \textbf{95\% CI} & \textbf{Sig.} \\
\midrule
DBSCAN & GMM & $-0.537$ & \textbf{$<$0.001} & $[-0.824, -0.250]$ & \textbf{Yes} \\
DBSCAN & K-Means & $-1.143$ & \textbf{$<$0.001} & $[-1.430, -0.856]$ & \textbf{Yes} \\
GMM & K-Means & $-0.606$ & \textbf{$<$0.001} & $[-0.893, -0.319]$ & \textbf{Yes} \\
\bottomrule
\end{tabular}
\end{table}

\begin{table}[htbp]
\centering
\caption{Tukey HSD: Polygon representation (Bengaluru). Not significant.}
\label{tab:supp_tukey_poly_blr}
\begin{tabular}{@{}llrrrc@{}}
\toprule
\textbf{Group 1} & \textbf{Group 2} & \textbf{Mean Diff.} & \textbf{$p$-adj} & \textbf{95\% CI} & \textbf{Sig.} \\
\midrule
Convex hull & Grid & $-0.210$ & 0.072 & $[-0.429, 0.009]$ & No \\
\bottomrule
\end{tabular}
\end{table}

%%%%%%%%%%%%%%%%%%%%%%%%%%%%%%%%%%%%%%%%%%%%%%%%%%%%%%%%%%%%%%%%%%%%%%%%%%%%%%%%%%%%%%%%%%%%%%%%%%%%
\section{Tukey HSD Pairwise Comparisons: UA-DETRAC Dataset}
\label{supp:tukey_uad}

Tables~\ref{tab:supp_tukey_roi_uad}--\ref{tab:supp_tukey_poly_uad} present the corresponding comparisons on the UA-DETRAC dataset, ordered by pipeline step number. On this dataset, all six parameters yield at least one significant pairwise difference.

\begin{table}[htbp]
\centering
\caption{Tukey HSD: ROI estimation (UA-DETRAC). Significant.}
\label{tab:supp_tukey_roi_uad}
\begin{tabular}{@{}llrrrc@{}}
\toprule
\textbf{Group 1} & \textbf{Group 2} & \textbf{Mean Diff.} & \textbf{$p$-adj} & \textbf{95\% CI} & \textbf{Sig.} \\
\midrule
Without ROI & With ROI & $-3.798$ & \textbf{$<$0.001} & $[-4.318, -3.278]$ & \textbf{Yes} \\
\bottomrule
\end{tabular}
\end{table}

\begin{table}[htbp]
\centering
\caption{Tukey HSD: Point selection, $s$ (UA-DETRAC). Significant differences ($p<0.05$) in bold.}
\label{tab:supp_tukey_points_uad}
\begin{tabular}{@{}llrrrc@{}}
\toprule
\textbf{Group 1} & \textbf{Group 2} & \textbf{Mean Diff.} & \textbf{$p$-adj} & \textbf{95\% CI} & \textbf{Sig.} \\
\midrule
$s=1$ & $s=3$ & $-2.920$ & \textbf{$<$0.001} & $[-4.232, -1.607]$ & \textbf{Yes} \\
$s=1$ & $s=5$ & $-3.670$ & \textbf{$<$0.001} & $[-4.982, -2.357]$ & \textbf{Yes} \\
$s=1$ & $s=7$ & $-4.737$ & \textbf{$<$0.001} & $[-6.049, -3.425]$ & \textbf{Yes} \\
$s=1$ & $s=9$ & $-4.793$ & \textbf{$<$0.001} & $[-6.105, -3.480]$ & \textbf{Yes} \\
$s=1$ & All & $-1.979$ & \textbf{$<$0.001} & $[-3.291, -0.667]$ & \textbf{Yes} \\
\midrule
$s=3$ & $s=5$ & $-0.750$ & 0.579 & $[-2.062, 0.562]$ & No \\
$s=3$ & $s=7$ & $-1.817$ & \textbf{0.001} & $[-3.130, -0.505]$ & \textbf{Yes} \\
$s=3$ & $s=9$ & $-1.873$ & \textbf{0.001} & $[-3.185, -0.561]$ & \textbf{Yes} \\
$s=3$ & All & $0.941$ & 0.318 & $[-0.372, 2.253]$ & No \\
\midrule
$s=5$ & $s=7$ & $-1.067$ & 0.187 & $[-2.380, 0.245]$ & No \\
$s=5$ & $s=9$ & $-1.123$ & 0.143 & $[-2.435, 0.189]$ & No \\
$s=5$ & All & $1.691$ & \textbf{0.003} & $[0.378, 3.003]$ & \textbf{Yes} \\
\midrule
$s=7$ & $s=9$ & $-0.056$ & 1.000 & $[-1.368, 1.257]$ & No \\
$s=7$ & All & $2.758$ & \textbf{$<$0.001} & $[1.446, 4.070]$ & \textbf{Yes} \\
\midrule
$s=9$ & All & $2.814$ & \textbf{$<$0.001} & $[1.501, 4.126]$ & \textbf{Yes} \\
\bottomrule
\end{tabular}
\end{table}

\begin{table}[htbp]
\centering
\caption{Tukey HSD: Density equalization (UA-DETRAC). Significant.}
\label{tab:supp_tukey_density_uad}
\begin{tabular}{@{}llrrrc@{}}
\toprule
\textbf{Group 1} & \textbf{Group 2} & \textbf{Mean Diff.} & \textbf{$p$-adj} & \textbf{95\% CI} & \textbf{Sig.} \\
\midrule
With subsampling & Without subsampling & $-2.204$ & \textbf{$<$0.001} & $[-2.727, -1.680]$ & \textbf{Yes} \\
\bottomrule
\end{tabular}
\end{table}

\begin{table}[htbp]
\centering
\caption{Tukey HSD: Exclusion zone ratio (UA-DETRAC). Significant differences ($p<0.05$) in bold.}
\label{tab:supp_tukey_excl_uad}
\begin{tabular}{@{}llrrrc@{}}
\toprule
\textbf{Group 1} & \textbf{Group 2} & \textbf{Mean Diff.} & \textbf{$p$-adj} & \textbf{95\% CI} & \textbf{Sig.} \\
\midrule
None & 1/3 & $-0.172$ & 1.000 & $[-2.333, 1.988]$ & No \\
None & 1/4 & $0.557$ & 0.989 & $[-1.604, 2.717]$ & No \\
None & 1/5 & $1.073$ & 0.766 & $[-1.088, 3.233]$ & No \\
None & 1/6 & $2.265$ & \textbf{0.033} & $[0.105, 4.425]$ & \textbf{Yes} \\
None & 1/7 & $4.044$ & \textbf{$<$0.001} & $[1.883, 6.204]$ & \textbf{Yes} \\
None & 1/8 & $7.828$ & \textbf{$<$0.001} & $[5.668, 9.988]$ & \textbf{Yes} \\
\midrule
1/3 & 1/4 & $0.729$ & 0.700 & $[-0.637, 2.095]$ & No \\
1/3 & 1/5 & $1.245$ & 0.102 & $[-0.122, 2.611]$ & No \\
1/3 & 1/6 & $2.437$ & \textbf{$<$0.001} & $[1.071, 3.803]$ & \textbf{Yes} \\
1/3 & 1/7 & $4.216$ & \textbf{$<$0.001} & $[2.849, 5.582]$ & \textbf{Yes} \\
1/3 & 1/8 & $8.000$ & \textbf{$<$0.001} & $[6.634, 9.367]$ & \textbf{Yes} \\
\midrule
1/4 & 1/5 & $0.516$ & 0.924 & $[-0.851, 1.882]$ & No \\
1/4 & 1/6 & $1.708$ & \textbf{0.004} & $[0.342, 3.074]$ & \textbf{Yes} \\
1/4 & 1/7 & $3.487$ & \textbf{$<$0.001} & $[2.121, 4.853]$ & \textbf{Yes} \\
1/4 & 1/8 & $7.271$ & \textbf{$<$0.001} & $[5.905, 8.638]$ & \textbf{Yes} \\
\midrule
1/5 & 1/6 & $1.192$ & 0.134 & $[-0.174, 2.559]$ & No \\
1/5 & 1/7 & $2.971$ & \textbf{$<$0.001} & $[1.605, 4.337]$ & \textbf{Yes} \\
1/5 & 1/8 & $6.756$ & \textbf{$<$0.001} & $[5.389, 8.122]$ & \textbf{Yes} \\
\midrule
1/6 & 1/7 & $1.779$ & \textbf{0.002} & $[0.412, 3.145]$ & \textbf{Yes} \\
1/6 & 1/8 & $5.563$ & \textbf{$<$0.001} & $[4.197, 6.930]$ & \textbf{Yes} \\
\midrule
1/7 & 1/8 & $3.785$ & \textbf{$<$0.001} & $[2.418, 5.151]$ & \textbf{Yes} \\
\bottomrule
\end{tabular}
\end{table}

\begin{table}[htbp]
\centering
\caption{Tukey HSD: Clustering algorithm (UA-DETRAC). All pairs are significant.}
\label{tab:supp_tukey_clust_uad}
\begin{tabular}{@{}llrrrc@{}}
\toprule
\textbf{Group 1} & \textbf{Group 2} & \textbf{Mean Diff.} & \textbf{$p$-adj} & \textbf{95\% CI} & \textbf{Sig.} \\
\midrule
DBSCAN & GMM & $-1.809$ & \textbf{$<$0.001} & $[-2.575, -1.043]$ & \textbf{Yes} \\
DBSCAN & K-Means & $-2.967$ & \textbf{$<$0.001} & $[-3.733, -2.201]$ & \textbf{Yes} \\
GMM & K-Means & $-1.158$ & \textbf{0.001} & $[-1.924, -0.392]$ & \textbf{Yes} \\
\bottomrule
\end{tabular}
\end{table}

\begin{table}[htbp]
\centering
\caption{Tukey HSD: Polygon representation (UA-DETRAC). Significant.}
\label{tab:supp_tukey_poly_uad}
\begin{tabular}{@{}llrrrc@{}}
\toprule
\textbf{Group 1} & \textbf{Group 2} & \textbf{Mean Diff.} & \textbf{$p$-adj} & \textbf{95\% CI} & \textbf{Sig.} \\
\midrule
Grid & Convex hull & $0.734$ & \textbf{0.006} & $[0.209, 1.260]$ & \textbf{Yes} \\
\bottomrule
\end{tabular}
\end{table}

%%%%%%%%%%%%%%%%%%%%%%%%%%%%%%%%%%%%%%%%%%%%%%%%%%%%%%%%%%%%%%%%%%%%%%%%%%%%%%%%%%%%%%%%%%%%%%%%%%%%
\section{Visual Examples: Effect of ROI and Exclusion Zone}
\label{supp:empirical}

\begin{figure*}[htbp]
    \centering
    \setlength{\tabcolsep}{3pt}
    \begin{tabular}{ccc}
    \textbf{No ROI, No EZ} & \textbf{EZ without ROI} & \textbf{EZ with ROI} \\[2pt]

    \includegraphics[width=0.32\textwidth]{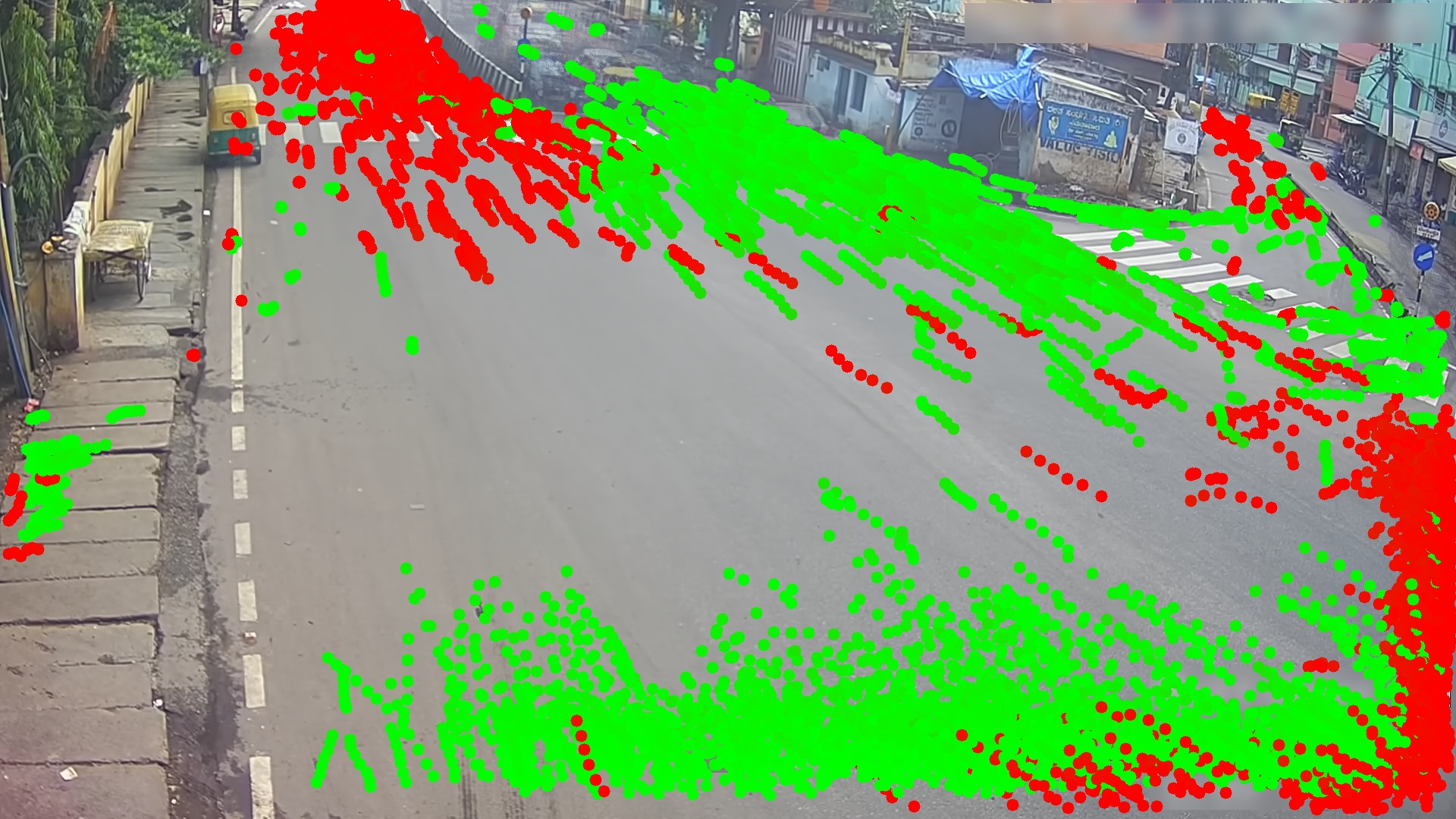} &
    \includegraphics[width=0.32\textwidth]{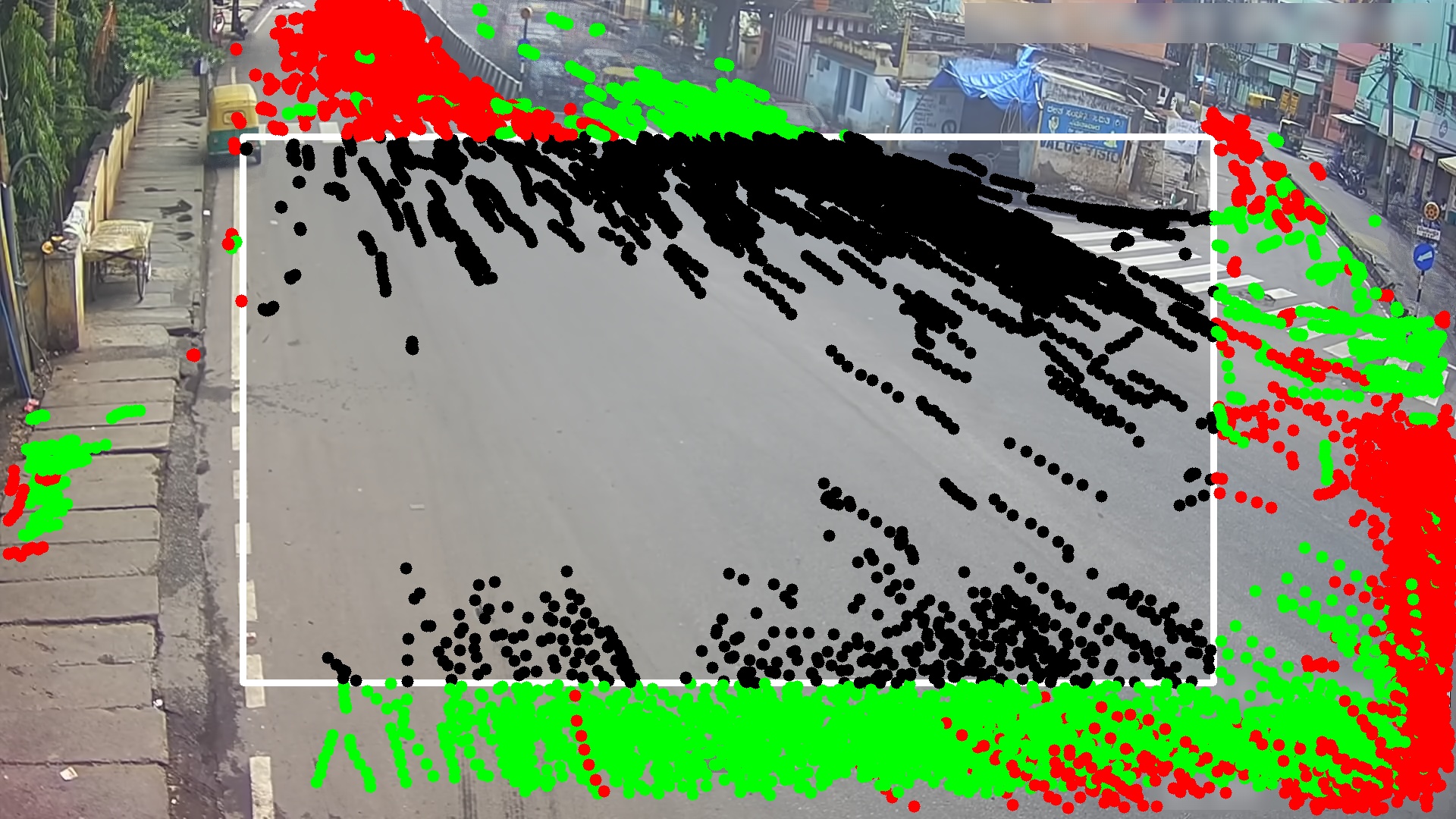} &
    \includegraphics[width=0.32\textwidth]{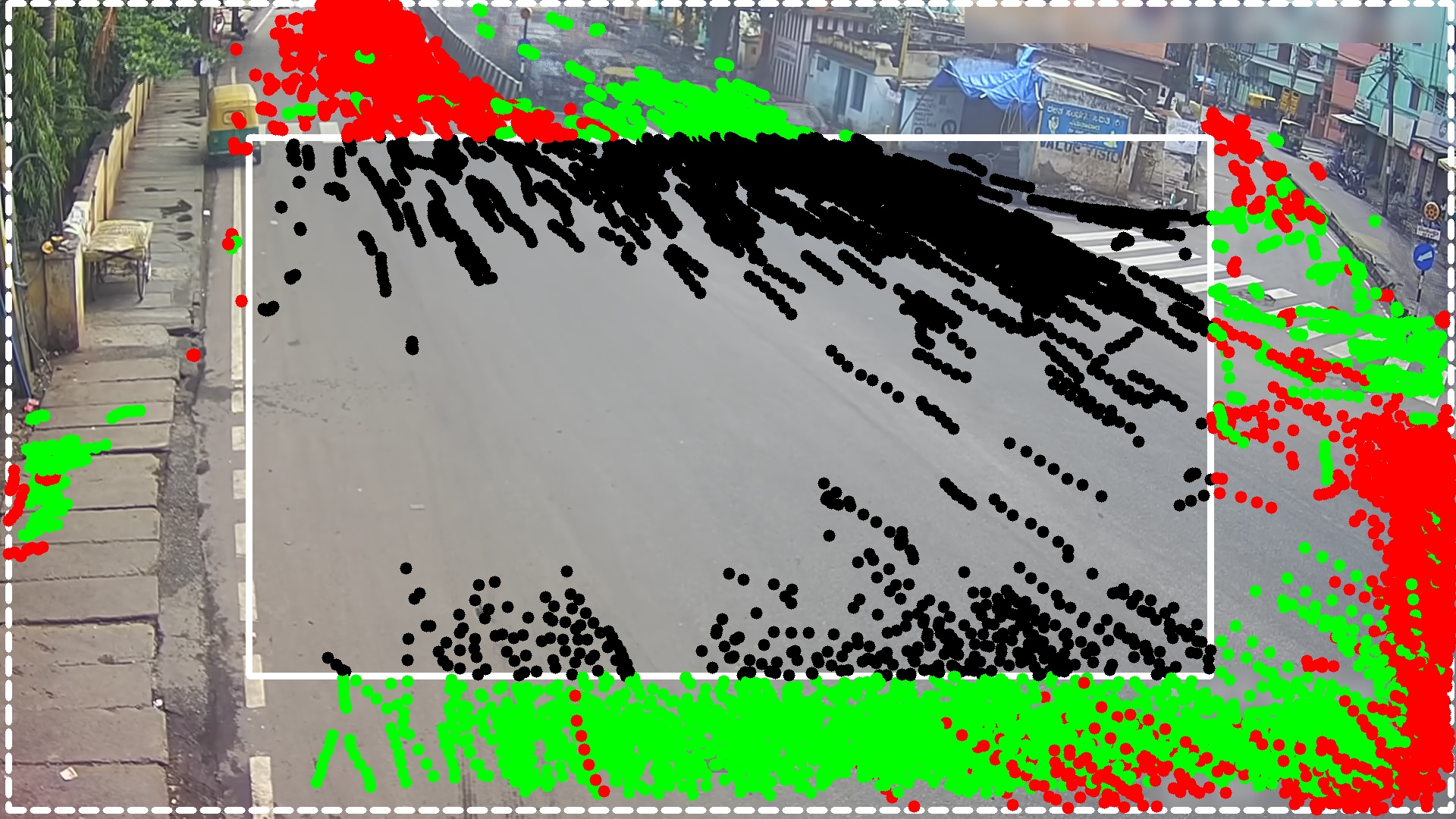} \\[4pt]

    \includegraphics[width=0.32\textwidth]{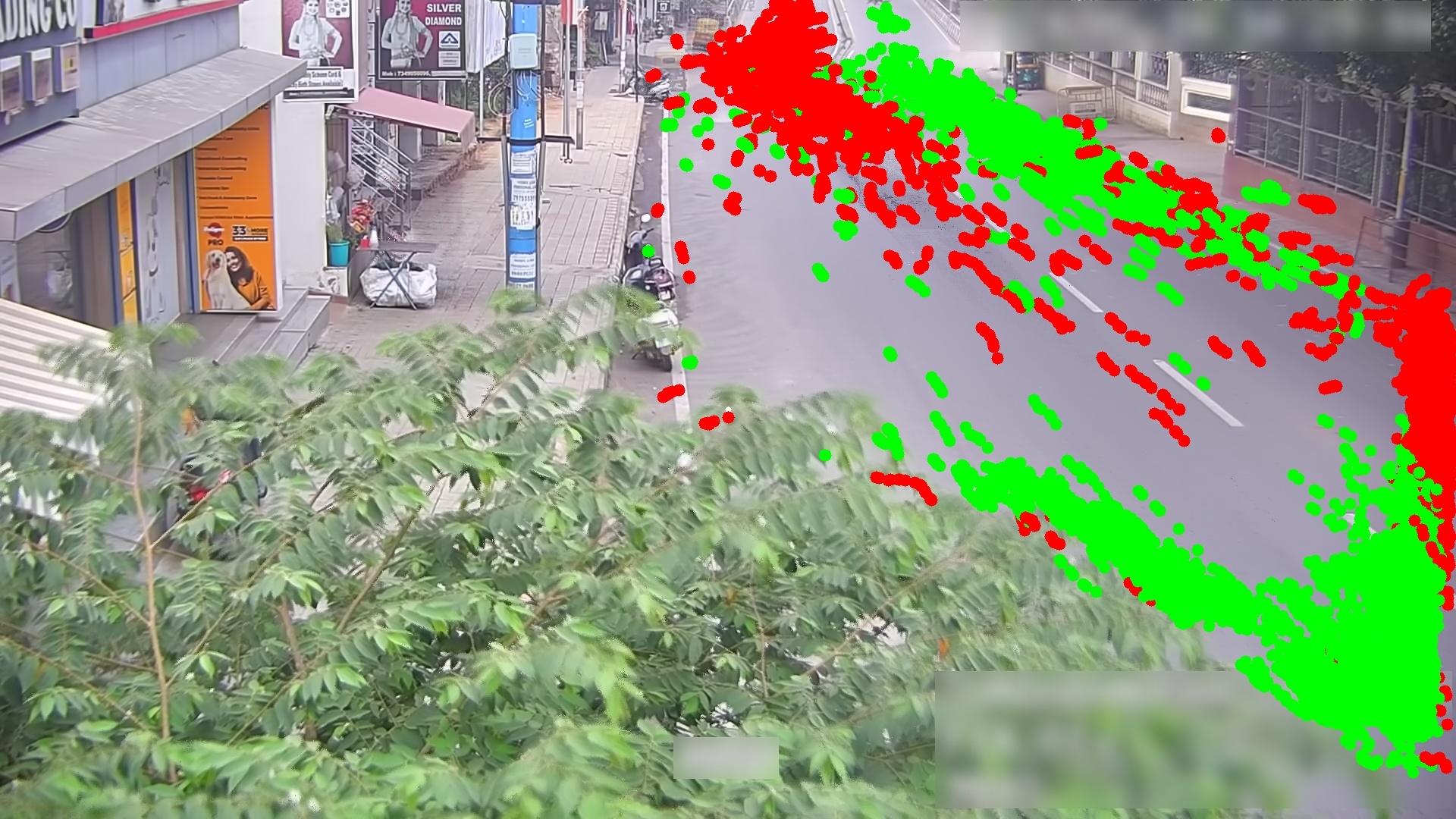} &
    \includegraphics[width=0.32\textwidth]{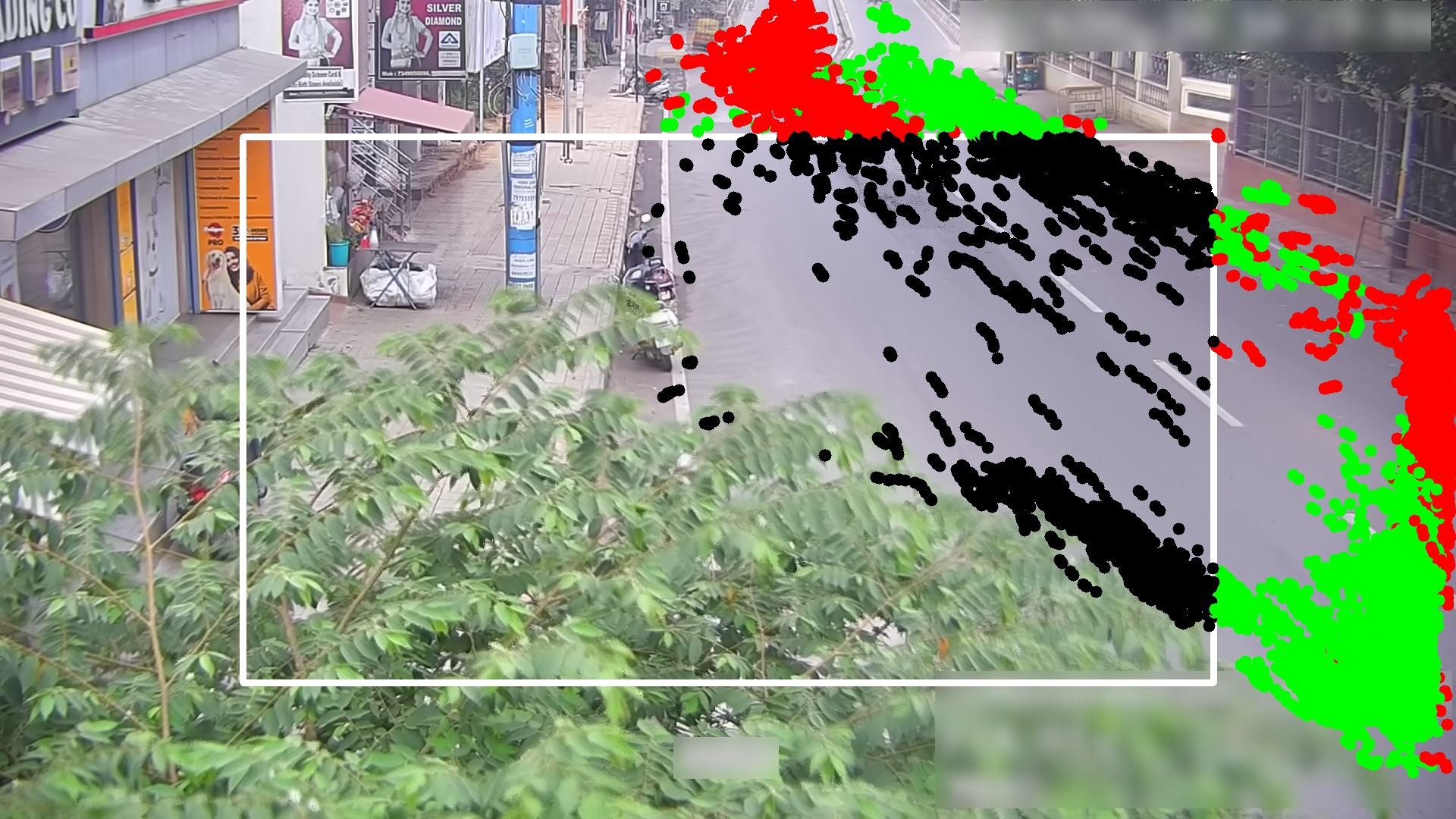} &
    \includegraphics[width=0.32\textwidth]{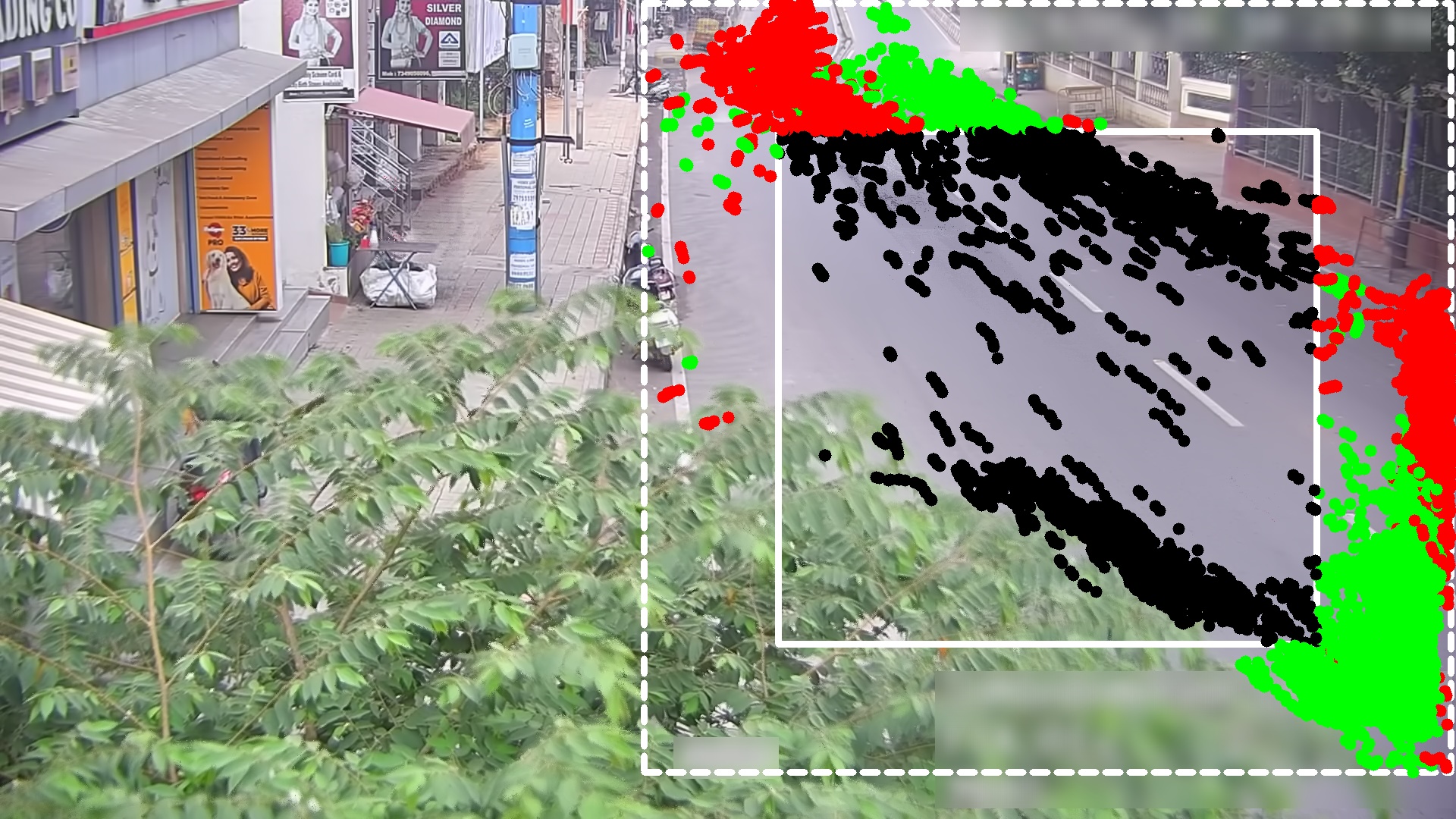} \\[4pt]

    \includegraphics[width=0.32\textwidth]{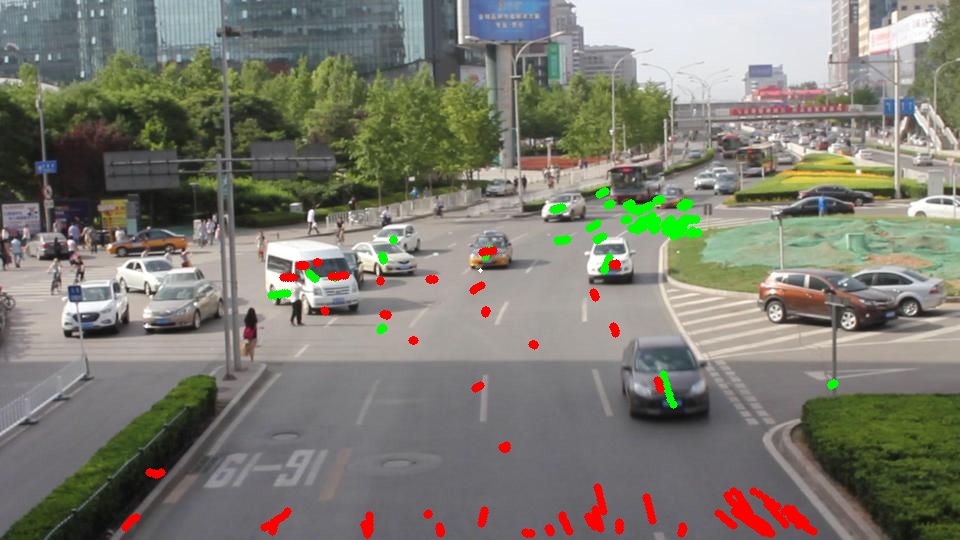} &
    \includegraphics[width=0.32\textwidth]{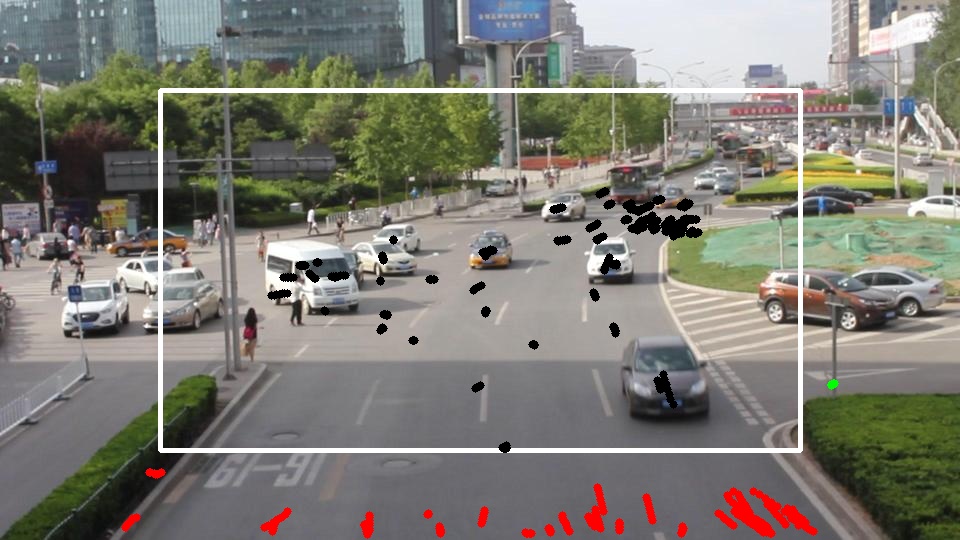} &
    \includegraphics[width=0.32\textwidth]{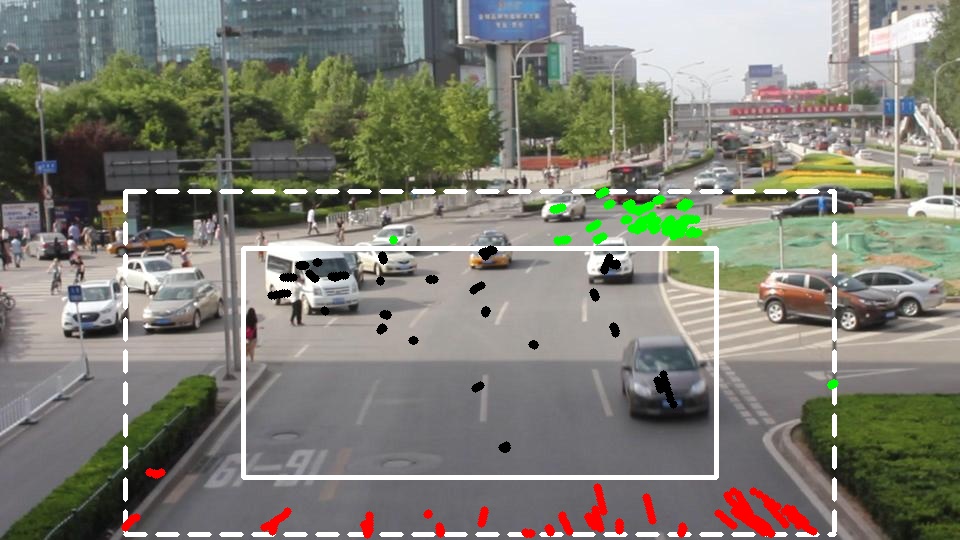} \\
    \end{tabular}
    \caption{Combined effect of ROI estimation and exclusion zone across three camera locations (top two rows: Bengaluru; bottom row: UA-DETRAC). Left: all selected points, no ROI or exclusion zone. Middle: exclusion zone applied without ROI estimation; points in the exclusion zone shown in black. Right: exclusion zone applied with ROI estimation. The effect of ROI estimation is most pronounced when the camera encompasses substantial non-road area (second and third rows), explaining the divergent statistical significance of ROI estimation across the two datasets (Tables~II and~III in the main text).}
    \label{fig:supp_roi_excl}
\end{figure*}

Fig.~\ref{fig:supp_roi_excl} illustrates the combined effect of ROI estimation (Step~2) and the exclusion zone (Step~4) across three camera locations: two from the Bengaluru dataset and one from UA-DETRAC. Each row corresponds to one location, and the three columns show progressive configurations: no ROI or exclusion zone (left), exclusion zone without ROI estimation (middle), and exclusion zone with ROI estimation (right). Points falling within the exclusion zone are shown in black.

The three rows exhibit a progression that mirrors the statistical findings in Section~\ref{sec:results}. In the first row, the camera tightly frames the road surface, so the trajectory bounding box approximates the full image frame; ROI estimation changes little, and the middle and right columns are nearly identical. In the second row, the camera includes some non-road area, and the exclusion zone without ROI estimation clips genuine entry points visible in the top-right portion of the frame; with ROI estimation, these points are preserved. The effect is most pronounced in the third row (UA-DETRAC), where the camera encompasses large non-road margins. Without ROI estimation, the exclusion zone is computed relative to the full image frame and removes genuine entry regions near the top of the scene entirely. With ROI estimation, the exclusion zone contracts to the trajectory-bounded area, preserving all entry and exit flows. This progression explains why ROI estimation is not statistically significant on the Bengaluru dataset, where cameras tend to frame the road tightly, but is highly significant on UA-DETRAC, where wider fields of view make the exclusion zone positioning consequential.

%%%%%%%%%%%%%%%%%%%%%%%%%%%%%%%%%%%%%%%%%%%%%%%%%%%%%%%%%%%%%%%%%%%%%%%%%%%%%%%%%%%%%%%%%%%%%%%%%%%%
\section{Camera View Diversity}
\label{supp:camera_diversity}

\begin{figure}[htbp]
    \centering
    \includegraphics[width=\textwidth]{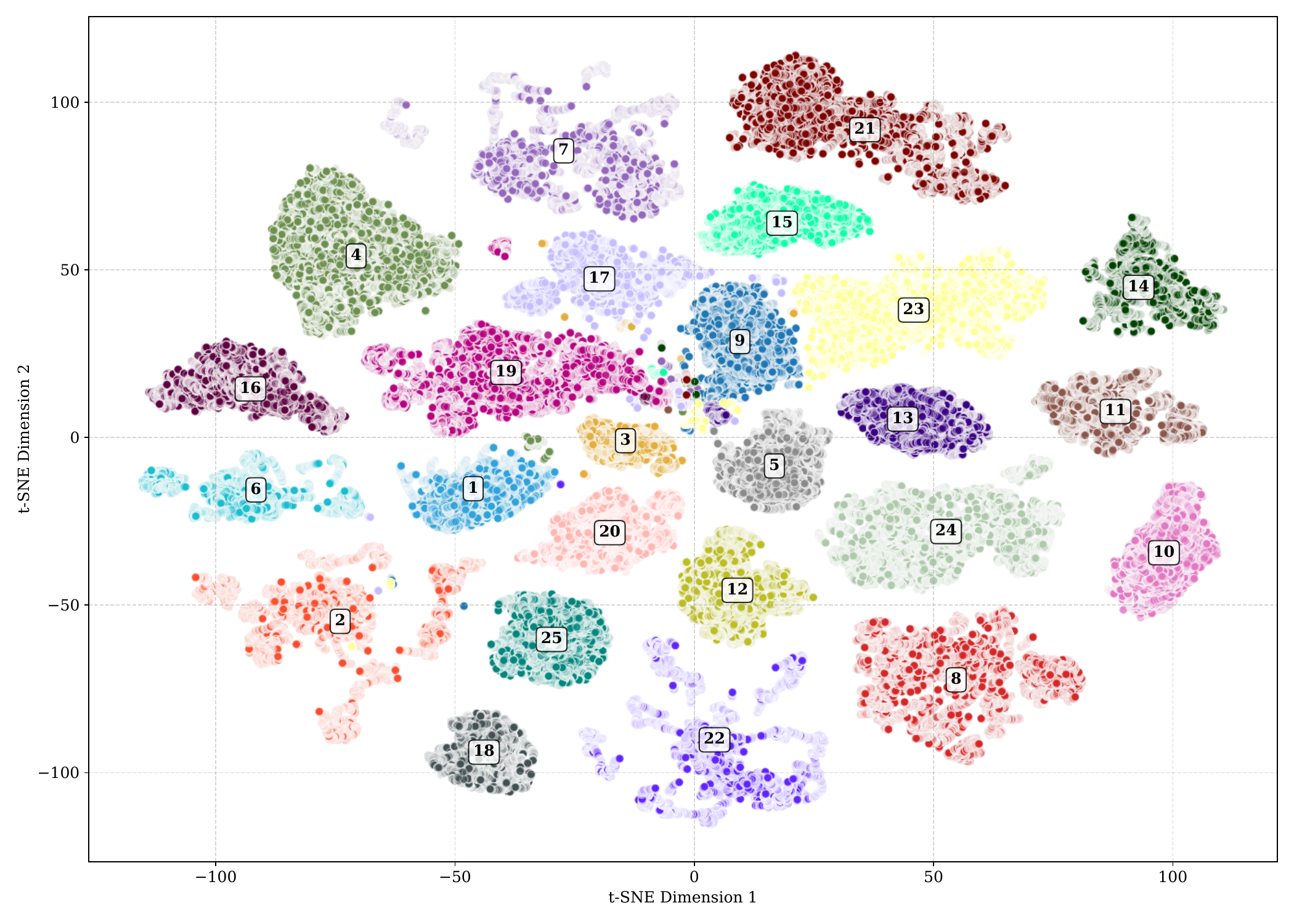}
    \caption{t-SNE visualization of the 25 Bengaluru camera locations. Each point represents a frame from the 15-minute clip at the respective location, embedded using DINOv3 (ViT-S/16). Distinct clusters with minimal overlap confirm the visual diversity of the selected camera feeds.}
    \label{fig:supp_tsne}
\end{figure}

To substantiate the diversity of the 25 Bengaluru camera locations, frames from each 15-minute clip were processed using a pretrained DINOv3 model~\cite{simeoni2025dinov3} with a ViT-S/16 backbone. The resulting 384-dimensional embeddings were reduced to two dimensions using t-SNE~\cite{van2008visualizing}. Fig.~\ref{fig:supp_tsne} shows that the 25 camera locations form distinct, well-separated clusters with minimal overlap, confirming that the selected cameras span a diverse range of visual scenes, viewing angles, and intersection geometries.

\end{document}